\documentclass[acmsmall]{acmart}

\usepackage{graphicx}
\usepackage{amsmath,multicol}
\usepackage[ruled,linesnumbered]{algorithm2e}
\usepackage{graphicx}
\usepackage{balance}

\usepackage{epstopdf}
\usepackage{multirow}
\usepackage{color,soul}

\usepackage{tabularx}
 
\usepackage[normalem]{ulem}
\usepackage{enumitem}

\usepackage{subcaption}

\usepackage{hyperref}

\renewcommand\vec[1]{\overrightarrow{#1}}
\newcommand\cev[1]{\overleftarrow{#1}}
\newcommand{\etal}{\textit{et~al. }}

\usepackage{url}

\newcolumntype{C}[1]{>{\centering\arraybackslash}p{#1}}
\AtBeginDocument{%
  \providecommand\BibTeX{{%
    \normalfont B\kern-0.5em{\scshape i\kern-0.25em b}\kern-0.8em\TeX}}}

\setcopyright{acmcopyright}
\copyrightyear{2018}
\acmYear{2018}
\acmDOI{10.1145/1122445.1122456}

\acmConference[ACM/IMS TDS]{ACM/IMS Transactions on Data Science}{New York, NY}{USA}
\acmBooktitle{ACM/IMS Transactions on Data Science}
\acmPrice{15.00}
\acmISBN{978-1-4503-XXXX-X/18/06}

\begin{document}

\title[Multi-Aspect Sentiment Analysis]{An Interpretable and Uncertainty Aware Multi-Task Framework for Multi-Aspect Sentiment Analysis}

\author{Tian Shi}
\email{tshi@vt.edu}
\affiliation{
  \institution{Virginia Tech}
  \country{USA}
}
\author{Ping Wang}
\email{ping@vt.edu}
\affiliation{
  \institution{Virginia Tech}
  \country{USA}
}
\author{Chandan K. Reddy}
\email{reddy@cs.vt.edu}
\affiliation{
  \institution{Virginia Tech}
  \country{USA}
}

\renewcommand{\shortauthors}{Shi et al.}

\begin{abstract}
In recent years, several online platforms have seen a rapid increase in the number of review systems that request users to provide aspect-level feedback. Document-level Multi-aspect Sentiment Classification (DMSC), where the goal is to predict the ratings/sentiment from a review at an individual aspect level, has become a challenging and an imminent problem. To tackle this challenge, we propose a deliberate self-attention based deep neural network model, named as FEDAR, for the DMSC problem, which can achieve competitive performance while also being able to interpret the predictions made. As opposed to the previous studies, which make use of hand-crafted keywords to determine aspects in rating predictions, our model does not suffer from human bias issues since aspect keywords are automatically detected through a self-attention mechanism. FEDAR is equipped with a highway word embedding layer to transfer knowledge from pre-trained word embeddings, an RNN encoder layer with output features enriched by pooling and factorization techniques, and a deliberate self-attention layer. In addition, we also propose an Attention-driven Keywords Ranking (AKR) method, which can automatically discover aspect keywords and aspect-level opinion keywords from the review corpus based on the attention weights. These keywords are significant for rating predictions by FEDAR.
Since crowdsourcing annotation can be an alternate way to recover missing ratings of reviews, we propose a LEcture-AuDience (LEAD) strategy to estimate model uncertainty in the context of multi-task learning, so that valuable human resources can focus on the most uncertain predictions.
Our extensive set of experiments on five different open-domain DMSC datasets demonstrate the superiority of the proposed FEDAR and LEAD models.
We further introduce two new DMSC datasets in the healthcare domain and benchmark different baseline models and our models on them.
Attention weights visualization results and  visualization of aspect and opinion keywords demonstrate the interpretability of our model and effectiveness of our AKR method.
\end{abstract}

\ccsdesc[500]{Information systems~Sentiment analysis}
\ccsdesc[500]{Information systems~Clustering and classification}
\ccsdesc[300]{Information systems~Information extraction}

\keywords{Multi-task learning, model uncertainty, deep neural network, dropout, classification, online reviews}

\maketitle

\section{Introduction}
\label{sec:introduction}

Sentiment analysis plays an important role in many business applications \cite{pang2008opinion}.
It is used to identify customers' opinions and emotions toward a particular product/service via identifying polarity (i.e., positive, neutral or negative) of given textual reviews \cite{liu2012sentiment,pang2002thumbs}.
In the past few years, with the rapid growth of online reviews, the topic of fine-grained aspect-based sentiment analysis (ABSA) \cite{pontiki2016semeval} has attracted significant attention since it allows models to predict opinion polarities with respect to aspect-specific terms in a sentence.
Different from sentence-level ABSA,
document-level multi-aspect sentiment classification (DMSC) aims to predict the sentiment polarity of documents, which are composed of several sentences, with respect to a given aspect \cite{yin2017document,li2018document,zeng2019variational}.
DMSC has become a significant challenge since many websites provide platforms for users to give aspect-level feedback and ratings, such as TripAdvisor\footnote{\url{https://www.tripadvisor.com}} and BeerAdvocate\footnote{\url{https://www.beeradvocate.com}}.
Fig.~\ref{fig:review-example} shows a review example from the BeerAdvocate website.
In this example, a beer is rated with four different aspects, i.e., feel, look, smell and taste.
The review also describes the beer with four different aspects.
There is an overall rating associated with this review.
Recent studies have found that users are less motivated to give aspect-level ratings \cite{yin2017document,zeng2019variational}, which makes it difficult to analyze their preference, and it takes a lot of time and effort for human experts to manually annotate them.

There are several recent studies that aim to predict the aspect ratings or opinion polarities using deep neural network based models with multi-task learning framework \cite{yin2017document,li2018document,zhang2019attentive,zeng2019variational}.
In this setting, rating predictions for different aspects, which are highly correlated and can share the same review encoder, are treated as different tasks. However, these models rely on hand-crafted aspect keywords to aid in rating/sentiment predictions \cite{yin2017document,li2018document,zhang2019attentive}.
Thus, their results, especially case studies of reviews, are biased towards pre-defined aspect keywords.
In addition, these models only focus on improving the prediction accuracy, however,
knowledge discovery (such as aspect and opinion related keywords) from review corpus still relies on unsupervised \cite{mcauley2012learning} and rule-based methods \cite{zeng2019variational}, which limits applications of current DMSC models \cite{yin2017document,li2018document,zhang2019attentive}.
In the past few years, model uncertainty of deep neural network classifiers has received increasing attention \cite{gal2016dropout,gal2016uncertainty}, because
it can identify low-confidence regions of input space and give more reliable predictions.
Uncertainty models have also been applied to deep neural networks for text classification \cite{zhang2019mitigating}.
However, few existing uncertainty methods have been used to improve the overall prediction accuracy of multi-task learning models when crowd-sourcing annotation is involved in the DMSC task.
In this paper, we attempt to tackle the above mentioned issues.
The primary contributions of this paper are as follows:

\begin{figure}[!t]
	\centering
	\includegraphics[width=0.6\textwidth]{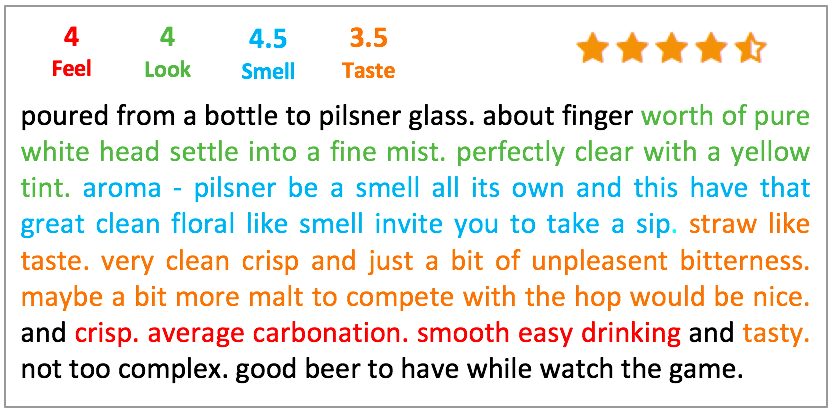}
	\caption{An example of an online review from BeerAdvocate platform. Keywords corresponding to different aspects are highlighted with different colors.}
	\label{fig:review-example}
\end{figure}

\begin{itemize}[leftmargin=*,topsep=1pt,itemsep=1pt, partopsep=1pt, parsep=1pt]
\item
Develop a FEDAR model that achieves competitive results on five benchmark datasets without using hand-crafted aspect keywords.
The proposed model is equipped with a highway word embedding layer, a sequential encoder layer whose output features are enriched by pooling and factorization techniques, and a deliberate self-attention layer.
The deliberate self-attention layer can boost performance as well as provide interpretability for our FEDAR model.
Here, FEDAR represents of some key components of our model, including Feature Enrichment, Deliberate self-Attention, and overall Rating.
\item
Introduce two new datasets obtained from the RateMDs website \url{https://www.ratemds.com}, which is a platform for patients to review the performance of their doctors.
We benchmark different models on them.
\item
Propose an Attention-driven Keywords Ranking (AKR) method to automatically discover aspect and opinion keywords from review corpus based on attention weights, which also provides a new research direction for interpreting self-attention mechanism.
The extracted keywords are significant to ratings/polarities predicted by FEDAR.
\item
Propose a LEcture-Audience (LEAD) method to measure the uncertainty of our FEDAR model for given reviews. 
This method can also be generally applied to other deep neural networks.
\end{itemize}

The rest of this paper is organized as follows:
In Section \ref{sec:related_work}, we introduce related work of the DMSC task and uncertainty estimation methods.
In Section \ref{sec:proposed_methods}, we present details of our proposed FEDAR model, AKR method and LEAD uncertainty estimation approach.
In Section \ref{sec:experiments}, we introduce different DMSC datasets, baseline methods and implementation details, as well as analyze experimental results.
Our discussion concludes in Section~\ref{sec:conclusion}.
\section{Related Work}
\label{sec:related_work}

Document-level Multi-aspect Sentiment Classification (DMSC) aims to predict ratings/sentiment of reviews with respect to given aspects.
It is originated from online review systems which request users to provide aspect-level ratings for a product or service.
Most of the early studies in DMSC solved this problem by first extracting features (e.g., $n$-grams) for each aspect and then predicting aspect-level ratings \cite{mcauley2012learning,lu2011multi} using regression techniques (e.g., Support Vector Regression \cite{smola2004tutorial}).
More recently, deep learning models formulate DMSC as a multi-task classification problem \cite{yin2017document,li2018document,zhang2019attentive}.
In these models, reviews are first encoded to their corresponding vector representation using recurrent neural networks.
Then, aspect-specific attention modules and classifiers are built upon the review encoders to predict the sentiment. For example, Yin~\etal  \cite{yin2017document} have formulated this task as a machine comprehension problem.
Li~\etal \cite{li2018document} proposed incorporating users' information, overall ratings, and hand-crafted aspect keywords into their model to predict ratings, instead of merely using textual reviews.
Zeng~\etal \cite{zeng2019variational}
introduced a variational approach to weakly supervised sentiment analysis.
Aspect-based sentiment classification (ABSA) \cite{pontiki2016semeval} is another research direction that is related to our work.
It consists of several fine-grained sentiment classification tasks, including aspect category detection and polarity, and aspect term extraction and polarity.
However, these tasks primarily focus on sentence-level sentiment classification \cite{wang2016attention,tang2016aspect}, and typically need human experts to annotate aspect terms, categories, and entities.
In this paper, we focus on the DMSC problem and our model is also based on a multi-task learning framework.
In addition, we place more emphasis on the model interpretability, automatic aspect and opinion keywords discovery, and uncertainty estimation.

Model uncertainty of deep neural networks (NNs) is another research topic related to this work.
Bayesian NNs, which learn a distribution over weights, have been studied extensively and achieved competitive results for measuring uncertainty \cite{blundell2015weight,neal2012bayesian,louizos2016structured}.
However, they are often difficult to implement and computationally expensive compared with standard deep NNs.
Gal and Ghahramani \cite{gal2015dropout} proposed using Monte Carlo dropout to estimate uncertainty by applying dropout \cite{srivastava2014dropout} at testing time, which can be interpreted as a Bayesian approximation of the Gaussian process \cite{rasmussen2003gaussian}.
This method has gain popularity in practice \cite{kendall2018multi,mcallister2017concrete} since it is simple to implement and computationally more efficient.
Recently, Zhang \etal \cite{zhang2019mitigating} applied dropout-based uncertainty estimation methods to text classification.
Our paper proposes a new method for estimating uncertainty for deep NNs and we use it to measure the uncertainty of our models.
\section{Proposed Methods}
\label{sec:proposed_methods}

\begin{figure*}[!t]
	\centering
	\includegraphics[width=\textwidth]{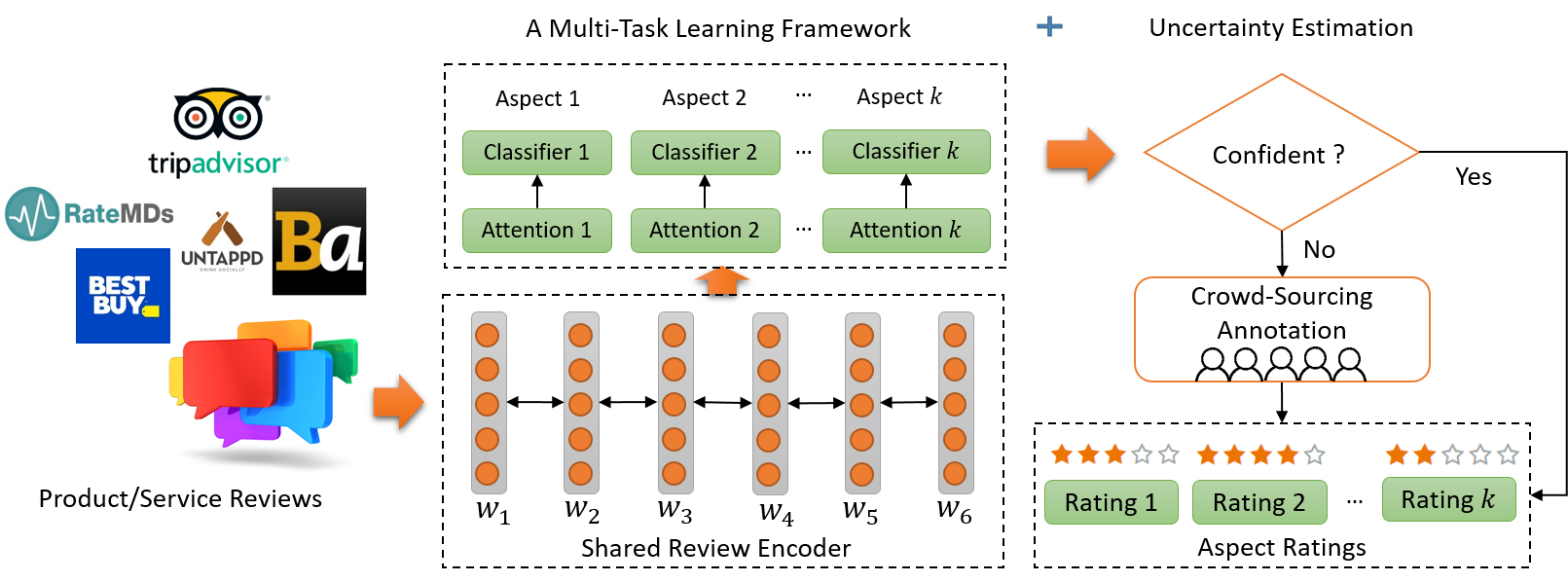}
	\caption{An overview of our multi-task learning framework with uncertainty estimation for accurate and reliable sentiment classification in DMSC task.
	Here, sentiment classification for each aspect is treated as a task and different tasks share the same review encoder.}
	\label{fig:mtArche}
\end{figure*}

\begin{figure}[!t]
	\centering
	\includegraphics[width=0.5\textwidth]{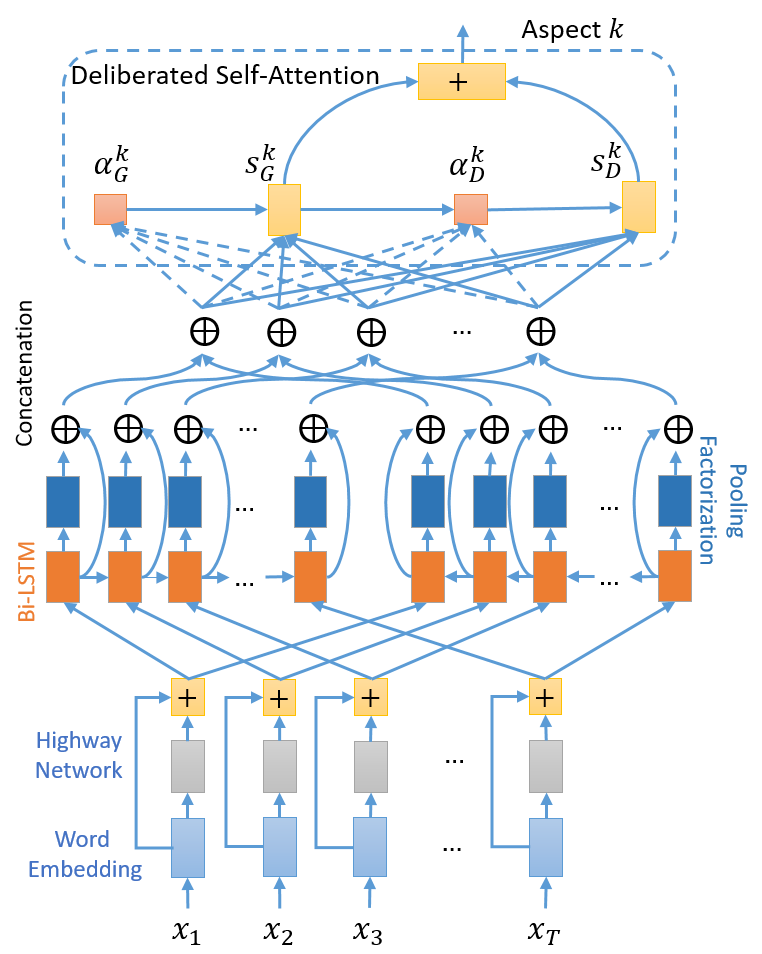}
	\caption{Review encoder and deliberate self-attention for aspect $k$. Each hidden state is enriched by three features, i.e., max-pooling, average-pooling and factorization.}
	\label{fig:model_struct}
\end{figure}

In this section, we first introduce our FEDAR model (See Fig.~\ref{fig:model_struct}) for the DMSC task.
Then, we describe our AKR method to automatically discover aspect and aspect-level sentiment terms based on the FEDAR model.
Finally, we discuss our LEAD method (See Fig.~\ref{fig:lead_method}) for measuring the uncertainty of the FEDAR model.

\subsection{The Proposed FEDAR Model}
\subsubsection{Problem Formulation}

The DMSC problem can be formulated as a multi-task classification problem, where the sentiment classification for each aspect is viewed as a task (See Fig.~\ref{fig:mtArche}). More formally, the DMSC problem is described as follows: 
Given a textual review $X=(x_1, x_2,...,x_T)$, our goal is to predict class labels, i.e., integer ratings/sentiment polarity of the review $y=(y^1, y^2, ..., y^K)$, where $T$ and $K$ are the number of tokens in the review and the number of aspects/tasks, respectively.
$x_t$ and $y^k$ are the one-hot vector representations of word $t$ and the class label of aspect $k$, respectively.

The challenge in this problem is to build a model that can achieve competitive accuracy without losing model interpretability or obtaining biased results.
Therefore, we propose improving word embedding, review encoder and self-attention layers to accomplish this goal.
We will now introduce our model and provide more details of our architecture in a layer-by-layer manner.

\subsubsection{Highway Word Embedding Layer}

This layer aims to learn word vectors based on pre-trained word embeddings.
We first use word embedding technique \cite{mikolov2013distributed} to map one-hot representations of tokens $x_1, x_2,...,x_T$ to a continuous vector space, thus, they are represented as $E_{x_1}, E_{x_2}, ..., E_{x_T}$, where $E_{x_t}$ is the word vector of $x_t$, pre-trained on a large corpus and fixed during parameter inference.
In our experiments, we adopted GloVe word vectors \cite{pennington2014glove}, so that they do not need to be trained from random states, which may result in poor embeddings due to the lack of word co-occurrence.

Then, a single layer highway network \cite{srivastava2015highway} is used to adapt the knowledge, i.e., semantic information from pre-trained word embeddings, to target DMSC datasets.
Formally, the highway network is defined as follows:
\begin{equation}
    E'_{x_t} = f(E_{x_t})\odot g(E_{x_t}) + E_{x_t}\odot (1-g(E_{x_t}))
\end{equation}
where $f(\cdot)$ and $g(\cdot)$ are affine transformations with ReLU and Sigmoid activation functions, respectively. $\odot$ represents element-wise product.
$g(\cdot)$ is also known as gate, which is used to control the information that is being carried to the next layer.
Intuitively, the highway network aims at transferring knowledge from pre-trained word embeddings to the target review corpus.
$E'_{x_t}$ can be viewed as a perturbation of $E_{x_t}$, and $f(\cdot)$ and $g(\cdot)$ have significantly fewer parameters than $E_{x_t}$. Therefore, training a highway network is more efficient than training a word embedding layer from random parameters.

\subsubsection{Review Encoder Layer}

This layer describes the review encoder and feature enrichment techniques proposed in our model.

\vspace{2mm}
\noindent\textbf{Sequential Encoder Layer:}
The output of highway word embedding layer ($E'_{x_1}, E'_{x_2}, ..., E'_{x_T}$) is fed into a sequential encoder layer.
Here, we adopt a multi-layer bi-directional LSTM encoder \cite{hochreiter1997long},
which encodes a review into a sequence of hidden states in forward direction $\vec{H}=(\vec{h_1}, \vec{h_2},...,\vec{h_T})$ and backward direction $\cev{H}=(\cev{h_1}, \cev{h_2},...,\cev{h_T})$.

\vspace{2mm}
\noindent\textbf{Representative Features:}
For each hidden state $\vec{h_t}$ (or $\cev{h_t}$), we generate three representative features, which will be later used to assist the attention mechanism to learn the overall review representation.

The first and second features, denoted by $\vec{h_t^\text{max}}$ and $\vec{h_t^\text{avg}}$, are the max-pooling and average-pooling of $\vec{h_t}$, respectively.
The third one is obtained using factorization machine \cite{rendle2010factorization}, where the factorization operation is defined as 
\begin{equation}
\label{eq:fm}
\mathcal{F}(z)=w_0+\sum_{i=1}^N w_iz_i+\sum_{i=1}^{N}\sum_{j=i+1}^{N}\left<V_i,V_j\right>z_iz_j.
\end{equation}
Here, the model parameters are $w_i\in\mathbb{R}$ and $V\in\mathbb{R}^{N\times F}$. $N$ and $F$ are the dimensions of the input vector $z$ and factorization, respectively.
$\left<\cdot,\cdot\right>$ is the dot product between two vectors.
$w_0$ in Eq. (\ref{eq:fm}) is a global bias, 
$w_i$ is the strength of the $i$-th variable, and
$\left<V_i,V_j\right>$ captures the pairwise interaction between $z_i$ and $z_j$.

Intuitively, the max-pooling and avg-pooling provide the approximated location (bound and mean) of the hidden state $\vec{h_t}$ in the $N$ dimensional space, while the factorization captures all single and pairwise interactions.
Together they provide the high-level knowledge of that hidden state.

\noindent\textbf{Feature Augmentation:}
Finally, the aggregated hidden state $h_t$ at time step $t$ is obtained by concatenating hidden states in both directions and all representative features, i.e.,
\begin{equation}
\aligned
\vec{h_t}&=\vec{h_t}\oplus\vec{h_t^\text{max}}\oplus\vec{h_t^\text{avg}}\oplus\mathcal{F}(\vec{h_t}),
\\
\cev{h_t}&=\cev{h_t}\oplus\cev{h_t^\text{max}}\oplus\cev{h_t^\text{avg}}\oplus\mathcal{F}(\cev{h_t}),\\
h_t&=\vec{h_t}\oplus\cev{h_t}.
\endaligned
\end{equation}
Thus, the review is encoded into a sequence of aggregated hidden states $H=(h_1,h_2,\dots,h_T)$.

\subsubsection{Deliberate Self-Attention Layer}

Once the aggregated hidden states for each review are obtained, we apply a self-attention layer for each task to learn an overall review representation for that task.
Compared with pooling and convolution operations, self-attention mechanism is more interpretable, since it can capture relatively important words for a given task.
However, a standard self-attention layer merely relies on a single global alignment vector across different reviews, which results in sub-optimal representations.
Therefore, we propose a deliberate self-attention alignment method to refine the review representations while maintaining the network interpretability.
In this section, we will first introduce the self-attention mechanism, and then provide the details of the deliberation counterpart.

\vspace{2mm}
\noindent\textbf{Global Self-Attention:}
For each aspect $k$, the self-attention mechanism \cite{yang2016hierarchical} is used to learn the relative importance of tokens in a review to the sentiment classification task.
Formally, given the aggregated hidden states $H$ for a review, the alignment score $u_{t,G}^k$ and attention weight $\alpha_{t,G}^k$ are calculated as follows:
\begin{equation}
u_{t,G}^k=(v^k_G)^\top\tanh(W_G^kh_t+b_G^k),\ 
\alpha_{t,G}^k=\frac{\exp(u_{t,G}^k)}{\sum_{\tau=1}^T\exp(u_{\tau,G}^k)},
\label{eqn:self_attn}
\end{equation}
where $W_G^k$, $v_G^k$ and $b_G^k$ are model parameters.
$G$ represents global, as the above attention mechanism is also known as global attention \cite{luong2015effective}. 
$v_\text{G}^k$ is viewed as a global aspect-specific base-vector in this paper, since it has been used in calculating the alignment with different hidden states across different reviews.
It can also be viewed as a global aspect-specific filter that is designed to capture important information for a certain aspect from different reviews.
Therefore, we also refer a regular self-attention layer as a global self-attention layer.
With attention weights, the global review representation is calculated by taking the weighted sum of all aggregated hidden states, i.e.,
$s^k_G=\sum_{t=1}^T\alpha_{t,G}^k h_t$.
Traditionally, $s^k_G$ is used for the sentiment classification task.

\vspace{2mm}
\noindent\textbf{Deliberate Attention:}
As we can see from Eq.~(\ref{eqn:self_attn}), the importance of a token $t$ is measured by the similarity between $\tanh(W_G^kh_t+b_G^k)$ and the base-vector $v_G^k$.
However, a single base-vector $v_G^k$ is difficult to capture the variability in the reviews, and hence, such alignment results in sub-optimal representations of reviews.
In this paper, we attempt to alleviate this problem by reusing the output of the global self-attention, i.e., $s^k_G$, as a document-level aspect-specific base-vector to produce better review representations.
Notably, $s^k_G$ already incorporates the knowledge of the review content and aspect $k$.
We refer this step as deliberation.

Given the hidden states $H$ and review representation $s^k_G$, we first calculate the alignment scores and attention weights as follows:
\begin{equation}
u_{t,D}^{k}=(s^{k}_G)^\top\tanh(W_D^k h_t+b_D^k),\ 
\alpha_{t,D}^k=\frac{\exp(u_{t,D}^k)}{\sum_{\tau=1}^T\exp(u_{\tau,D}^k)},
\label{eqn:deli_attn}
\end{equation}
where $W_D^k$ and $b_D^k$ are parameters. $D$ represents deliberation. 
Similarly, we can calculate the aspect-specific review representation by deliberation as
$s^k_D=\sum_{t=1}^T\alpha_{t,D}^k h_t$.

\vspace{2mm}
\noindent\textbf{Review Representation:}
Finally, the review representation for aspect $k$ can be obtained as follows\footnote{In this paper, we also consider models that repeat the deliberation for multiple times. However, we did not observe significant performance improvement.}:
\begin{equation}
    s^k=s^k_G + s^k_D
    = \sum_{t=1}^T \big(\alpha^k_{t,G} + \alpha^k_{t,D}\big)h_t.
    \label{eqn:deli-context}
\end{equation}
From the above equation, we not only get refined review representations but also maintain the interpretability of our model.
Here, we did not use the concatenation of two vectors since we would like to maintain the interpretability as well.
Notably, we can use the accumulated attention weights, i.e., $\frac{1}{2}(\alpha^k_{t,G} + \alpha^k_{t,D})$, to interpret our experimental results.

\subsubsection{Sentiment Classification Layer}

Finally, we pass the representation of each review for aspect $k$ into an aspect-specific classifier to get the probability distribution over different class labels.
Here, the classifier is defined as a two layer feed-forward network with a ReLU activation followed by a softmax layer, i.e.,
\begin{equation}
\aligned
y^k_\text{out}&=\text{ReLU}(W^k_\text{out}s^k+b_\text{out}^k), \\
y^k_\text{pred}&=\text{softmax}(W_\text{pred}^ky^k_\text{out}+b_\text{pred}^k),
\endaligned
\label{eqn:class_label_ditribution}
\end{equation}
where $W^k_\text{out}$, $W_\text{pred}^k$, $b_\text{out}^k$, and $b_\text{pred}^k$ are learnable parameters.

Given the ground-truth labels $\hat{y}^k$, which is a one-hot vector, our goal is to minimize the averaged cross-entropy error between $y^k_\text{pred}$ and $\hat{y}^k$ across all aspects, i.e., 
\begin{equation}
    \mathcal{L}_\theta=-\sum_{k=1}^K\sum_{i=1}^N \hat{y}^k_i\log(y^k_{\text{pred},i}),
\end{equation}
where $K$ and $N$ represents the number of aspects and class labels, respectively.
The model is trained in an end-to-end manner using back-propagation.

\subsection{Aspect and Sentiment Keywords}

Traditionally, aspect and sentiment keywords are obtained using unsupervised clustering methods, such as topic models \cite{mcauley2012learning,shi2019document}.
However, these methods cannot automatically build correlations between keywords and aspects or sentiment due to the lack of supervision.
Aspect and opinion term extractions in fine-grained aspect-based sentiment analysis tasks \cite{pontiki2014semeval,pontiki2016semeval,fan2019target,wang2017coupled} focus on extracting terms and phrases from sentences.
However, they require a number of labeled reviews to train deep learning models.
In this paper, we propose a fully automatic Attention-driven Keywords Ranking (AKR) method to discover aspect and opinion keywords, which are important to predicted ratings, from a review corpus based on self-attention (or deliberate self-attention) mechanism in the context of DMSC.

\subsubsection{Aspect Keywords Ranking}

The significance of a word $w$ to an aspect $k$ can be described by a conditional probability $p_\mathcal{C}(w|k)$ on a review corpus~$\mathcal{C}$.
Intuitively, given an aspect $k$, if a word $w_1$ is more frequent than $w_2$ across the corpus, then, $w_1$ is more significant to aspect $k$.
We can further expand this probability as follows:
\begin{equation}
    p_\mathcal{C}(w|k)= \sum_{\xi\in\mathcal{C}} p_\mathcal{C}(w,\xi|k),
    \label{eqn:keywords_expansion}
\end{equation}
where $\xi$ is a review in corpus $\mathcal{C}$.
For each $\xi\in\mathcal{C}$, probability $p_\mathcal{C}(w,\xi|k)$ indicates the importance of word $w$ to the aspect $k$, which can be defined using attention weights, i.e.,
\begin{equation}
    p_\mathcal{C}(w,\xi|k)=\frac{\sum_{t=1}^T\alpha^\xi_t\cdot\delta(w_t,w)}{\sum_{\xi'\in\mathcal{C}}f_{\xi'}(w)+\gamma},
    \label{eqn:keywords_attention_def}
\end{equation}
where $f_{\xi'}(w)$ is frequency of $w$ in document $\xi'$ and $\gamma$ is a smooth factor.
$\delta(w_t,w)=\begin{cases}1& \text{if } w_t=w\\ 0& \text{otherwise}\end{cases}$ 
is a delta function.
Attention weight $\alpha_t^\xi$ is defined as  $\alpha_t^\xi=\frac{1}{2}(\alpha^k_{t,G} + \alpha^k_{t,D})$ for the deliberation self-attention mechanism.
In Eq.~(\ref{eqn:keywords_attention_def}), the denominator is applied to reduce the noise from stop-words and punctuation.
After obtaining the score $p_\mathcal{C}(w|k)$ for every member in the vocabulary, we collect top-ranked words (with part-of-speech tags: NOUN and PROPN) as aspect keywords.

\subsubsection{Aspect-level Opinion Keywords}

Similarly, we can estimate the significance of a word $w$ to an aspect-level opinion label/rating $\hat{y}^k$ by a conditional probability $p_\mathcal{C}(w|\hat{y}^k)$.
Let us use $\mathcal{C}_{\hat{y}^k}$ to denote reviews with rating $\hat{y}^k$ for aspect $k$, then, the following equivalence holds, i.e.,
\begin{equation}
    p_\mathcal{C}(w|\hat{y}^k)=p_{\mathcal{C}_{\hat{y}^k}}(w|k),
    \label{eqn:sentiment_keywords}
\end{equation}
which can be further calculated by Eqs.~(\ref{eqn:keywords_expansion}) and (\ref{eqn:keywords_attention_def}).
Intuitively, we first construct a subset $\mathcal{C}_{\hat{y}^k}\subset\mathcal{C}$ of the review corpus, then, we use attention weights of aspect $k$ to calculate the significance of word $w$ to that aspect.
Finally, we collect top-ranked words (with part-of-speech tags: ADJ, ADV and VERB) as aspect-level opinion keywords.

\begin{figure}[!tp]
	\centering
	\includegraphics[width=0.5 \textwidth]{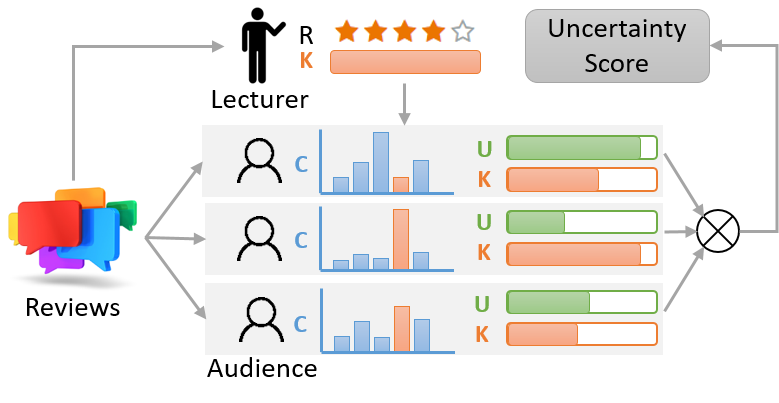}
	\caption{The LEcture-AuDience (LEAD) model for uncertainty estimation. `R', `K', `C', `U' represent rating, knowledge, probability distribution of different class labels (see Eq.~(\ref{eqn:class_label_ditribution})), and uncertainty score, respectively.}
	\label{fig:lead_method}
\end{figure}

\subsection{The Proposed Uncertainty Model}

Although our FEDAR model has achieved competitive prediction accuracy and our AKR method allows us to explore aspect and sentiment keywords, it is still difficult to deploy such a model in real-world applications.
In DMSC datasets, we find that there are many typos and abbreviations in reviews and many reviews describe the product or service from only one aspect.
However, deep learning models cannot capture these problems in the datasets, therefore, the predictions are not reliable.
One way to tackle this challenge is by estimating the uncertainty of model predictions.
If a model returns ratings with high uncertainty, we can pass the review to human experts for annotation.
In this section, we propose a LEcture-AuDience (LEAD) method (See Fig.\ref{fig:lead_method}) to measure the uncertainty of our FEDAR model in the context of multi-task learning.

\subsubsection{Lecturer and Audiences}
We use a lecturer (denoted by $\mathcal{M}^L$) to represent any well-trained deep learning model, e.g., FEDAR model.
Audiences are models (denoted by $\mathcal{M}^A$) with partial knowledge of the lecturer, where \textit{knowledge can be interpreted as relationships between an input review and output ratings} which are inferred by $\mathcal{M}^L$.
Here, $\mathcal{M}^A=\{\mathcal{M}^{A_1}, \mathcal{M}^{A_2},...,\mathcal{M}^{A_{|A|}}\}$, where $|A|$ is the number of audiences.
\textit{Partial knowledge determines the eligibility of audiences to provide uncertainty scores.}
For example, eligible audiences can be: 
(1) Models obtained by pruning some edges (e.g., dropout with small dropout rate) of the lecturer model. 
(2) Models obtained by continuing training the lecturer model with very small learning rate for a few batches.
Ineligible audiences include:
(1) Random models trained on the same or a different review corpus.
(2) Models with the same or similar structure as lecturer but initialized with different parameters and trained on a different corpus.

\subsubsection{Uncertainty Scores}

Given a review, suppose the lecturer $\mathcal{M}^L$ predicts the class label as $\tilde{y}^{L,k}$ for aspect $k$, where $\tilde{y}^{L,k}$ is an one-hot vector.
An audience $\mathcal{M}^{A_\mu}$ obtains the probability distribution over different class labels as $y^{A_\mu,k}_\text{pred}$ (See Eq.~(\ref{eqn:class_label_ditribution})).
Then, the uncertainty score is defined as the cross entropy between $\tilde{y}^{L,k}$ and $y^{A_\mu,k}_\text{pred}$, which is calculated by
\begin{equation}
    \psi^{A_\mu, k}=-\sum_{i=1}^{N}\tilde{y}^{L,k}_i\log(y^{A_\mu,k}_{\text{pred},i}).
\end{equation}
\textit{Intuitively, the audience is more uncertain about the lecturer's prediction if it gets lower probability for that prediction.}
For example, in Fig.~\ref{fig:lead_method}, the lecturer model predicts rating/label as 4.
Three audiences obtain probability 0.1, 0.8, 0.5 for that label, respectively.
Then, their uncertainty scores are $\psi^{A_1,k}=2.30$, $\psi^{A_2,k}=0.22$, and $\psi^{A_3,k}=0.69$.

With the uncertainty score from a single audience and for a single aspect, we can calculate the final uncertainty score as
\begin{equation}
    \psi=\exp\sum_{\mu=1}^{|A|}\zeta\log\left(\exp{\sum_{k=1}^k\log\big(\psi^{A_\mu,k}+\lambda\big)}+\eta\right),
\end{equation}
where $\lambda$ and $\eta\geq 1$ are smoothing factor and set to $1$ in our experiments.
$\zeta$ is an empirical factor for knowledge.
If audience networks are obtained by applying dropout to the lecturer network, the higher the dropout rate, the lower the factor $\zeta$.
In this case, the audiences have less knowledge to the lecturer.

After obtaining uncertainty scores for all reviews in the testing set, we can select either certain percent of reviews with higher scores or reviews with scores over a threshold for crowdsourcing annotation.
Human experts are expected to analyze the reviews and decide the aspect ratings for them.


\section{Experimental Results}
\label{sec:experiments}

In this section, we present the results from an extensive set of experiments and demonstrate the effectiveness of our proposed FEDAR model, AKR and LEAD methods.

\subsection{Research Questions}

Our empirical analysis aims at the following Research Questions (RQs):
\begin{itemize}[leftmargin=*,topsep=1pt,itemsep=1pt, partopsep=1pt, parsep=1pt]
    \item[$\bullet$] \textbf{RQ1}: What is the overall performance of FEDAR? Does it outperform state-of-the-art baselines?
    \item[$\bullet$] \textbf{RQ2}: What is the overall performance of LEAD method compared with uncertainty estimation baselines? 
    \item[$\bullet$] \textbf{RQ3}: How does each component in FEDAR contribute to the overall performance?
    \item[$\bullet$] \textbf{RQ4}: Is the deliberate self-attention module interpretable? Does it learn meaningful aspect and opinion terms from a review corpus?
\end{itemize}

\subsection{Datasets}

\begin{table}[!t]
	\centering
	\caption{Statistics of different DMSC datasets. $\dagger$ indicates the datasets collected and prepared by us.}
	\begin{tabular}{|l|c|c|c|}
		\hline
        \bf Dataset & \bf \# docs & \bf \# aspects & \bf Scale
        \\\hline
        TripAdvisor-R & 29,391 & 7 & 1-5 \\\hline
        TripAdvisor-RU & 58,632 & 7 & 1-5 \\\hline
        TripAdvisor-B & 28,543 & 7 &1-2 \\\hline
        BeerAdvocate-R & 50,000 & 4 & 1-10\\\hline
        BeerAdvocate-B & 27,583 & 4 & 1-2\\\hline
        RateMDs-R$\dagger$ & 155,995 & 4 &  1-5\\\hline
        RateMDs-B$\dagger$ & 120,303 & 4 & 1-2
        \\
        \hline
	\end{tabular}
	\label{tab:dataset}
\end{table}

We first conduct our experiments on
five benchmark datasets, which are obtained from TripAdvisor and BeerAdvocate review platforms.
TripAdvisor based datasets have seven aspects (\textit{value, room, location, cleanliness, check in/front desk, service,
and business service}), while BeerAdvocate based datasets have four aspects (\textit{feel, look,
smell, and taste}).
TripAdvisor-R \cite{yin2017document}, TripAdvisor-U \cite{li2018document} and BeerAdvocate-R \cite{yin2017document,lei2016rationalizing} use the original rating scores as sentiment class labels.
In TripAdvisor-B and BeerAdvocate-B \cite{zeng2019variational}, the original scale is converted to a binary scale, where $1$ and $2$ correspond to negative and positive sentiment, respectively.
Neutral has been ignored in both datasets.
All datasets have been tokenized and split into train/development/test sets with a proportion of 8:1:1.
In our experiments, we use the same datasets that are provided by the previous studies in the literature \cite{yin2017document,li2018document,zeng2019variational}.
Statistics of the datasets are summarized in Table~\ref{tab:dataset}.

In addition to the aforementioned five datasets, we also propose two new datasets, i.e., RateMDs-R and RateMDs-B, and benchmarked our models on them.
RateMDs dataset was collected from \url{https://www.ratemds.com} website which has textual reviews along with numeric ratings for medical experts primarily in the North America region.
Each review comes with ratings of four different aspects, i.e., \textit{staff, punctuality, helpfulness, and knowledge}.
The overall rating is the average of these aspect ratings. 
To obtain a more refined dataset for our experiments, we removed reviews with missing aspect ratings and selected the rest of the reviews whose lengths are between 72 and 250 tokens (outliers \footnote{The average number of tokens for all reviews is 72 tokens and there are very few reviews with more than 250 tokens.}), since short reviews may not have information on all the four aspects.
The original data has a rating-imbalance problem, i.e., $60\%$ and $17\%$ of reviews are rated as 5 and 1, respectively, and more than $50\%$ of reviews have identical aspect ratings.
Therefore, similar to \cite{lei2016rationalizing},
we chose reviews with different aspect ratings, i.e., at least three of aspect ratings are different.
The statistics of our dataset is shown in Table~\ref{tab:dataset}.
For RateMDs-R, we tokenized reviews with Stanford corenlp\footnote{\url{https://stanfordnlp.github.io/CoreNLP/}} and randomly split the dataset into training, development and testing by a proportion of 135,995/10,000/10,000. For RateMDs-B, we followed the process in \cite{zeng2019variational} by  converting original scales to binary and sampling data according to the overall polarities to avoid the imbalance issue.
The statistics of the RateMDs-B dataset is also shown in Table~\ref{tab:dataset}.
Similarly, we split the dataset into training, development and testing by a proportion of 100,303/10,000/10,000.

\subsection{Comparison Methods}

To demonstrate the effectiveness of our methods, we compare the proposed models with following baseline methods:
\begin{itemize}[leftmargin=*,topsep=1pt,itemsep=1pt, partopsep=1pt, parsep=1pt]
    \item[$\bullet$] \textbf{MAJOR} simply uses the majority sentiment labels or polarities in training data as predictions.
    \item[$\bullet$] {\bf GLVL} first calculates the document representation by averaging the word vectors of all keywords in a review, where pre-trained word vectors are obtained from GloVe \cite{pennington2014glove}. Then, a LIBLINEAR package \cite{fan2008liblinear} is used for the classification task.
    \item[$\bullet$] {\bf BOWL} feeds normalized Bag-of-Words (BOW) representation of reviews into the LIBLINEAR package for the sentiment classification. In our experiments, stop-words and punctuation are removed in order to enable the model to capture the keywords more efficiently.
    \item[$\bullet$] \textbf{MCNN} is an extension of the CNN model in the multi-task learning framework. For each task, CNN \cite{kim2014convolutional} extracts key features from a review by applying convolution and max-over-time pooling \cite{collobert2011natural} operations over the shared word embeddings layer.
    \item[$\bullet$] \textbf{MLSTM} extends a multi-layer Bi-LSTM model \cite{hochreiter1997long}, which captures both forward and backward semantic information, with the multi-task learning framework, where different tasks have their own classifiers and share the same Bi-LSTM encoder.
    \item[$\bullet$] \textbf{MBERT} is a multi-task version of the BERT classification model \cite{devlin2019bert}. Different tasks share the same BERT encoder \cite{Wolf2019HuggingFacesTS}.
    \item[$\bullet$] \textbf{MATTN} is a multi-task version of self-attention based models. Similar to MLSTM, different tasks share the same Bi-LSTM encoder.
    For each task, we first apply a self-attention layer, and then pass the document representations to a sentiment classifier.
    \item[$\bullet$] \textbf{DMSCMC} \cite{yin2017document} introduces a hierarchical iterative attention model to build aspect-specific document representations by frequent and repeated interactions between documents and aspect questions.
    \item[$\bullet$] \textbf{HRAN} \cite{li2018document} incorporates hand-crafted aspect keywords and the overall rating into a hierarchical network to build sentence and document representations.
    \item[$\bullet$] \textbf{AMN} \cite{zhang2019attentive} first uses attention-based memory networks to incorporate hand-crafted aspect keywords information into the aspect and sentence memories. Then, recurrent attention operation and multi-hop attention memory networks are employed to build document representations.
    \item[$\bullet$] \textbf{FEDAR} We name our model as {FEDAR}, where FE, DA and R represent Feature Enrichment, Deliberate self-Attention, and overall Rating, respectively.
\end{itemize}
\vspace{0.1in}
We compare our LEAD method with the following uncertainty estimation approaches:
\begin{itemize}[leftmargin=*,topsep=1pt,itemsep=1pt, partopsep=1pt, parsep=1pt]
    \item[$\bullet$] \textbf{Max-Margin} is the maximal activation of the sentiment classification layer (after softmax normalization).
    \item[$\bullet$] \textbf{PL-Variance} (Penultimate Layer Variance) \cite{zaragoza1998confidence} uses the variance of the output of the sentiment classification layer (before softmax normalization) as the uncertainty score.
    \item[$\bullet$] \textbf{Dropout} \cite{gal2015dropout} apply dropout to deep neural networks during training and testing. The dropout can be used as an approximation of Bayesian inference in deep Gaussian processes, which aims to identify low-confidence regions of input space.
\end{itemize}
All methods are based on our FEDAR model.

\subsection{Implementation Details} 

We implemented all deep learning models using Pytorch \cite{paszke2017automatic} and the best set of parameters are selected based on the development set.
Word embeddings are pre-loaded with 300-dimensional GloVe embeddings \cite{pennington2014glove} and fixed during training.
For MCNN, filter sizes are chosen to be 3, 4, 5 and the number of filters are 400 for each size.
For all LSTM based models, the dimension of hidden states is set to 600 and the number of layers is 4.
All parameters are trained using ADAM optimizer \cite{kingma2014adam} with an initial learning rate of 0.0005.
The learning rate decays by 0.8 every 2 epochs.
Dropout with a dropout-rate 0.2 is applied to the classifiers. 
Gradient clipping with a threshold of 2 is also applied to prevent gradient explosion.
For MBERT, we leveraged the pre-trained BERT encoder from HuggingFace's Transformers package  \cite{Wolf2019HuggingFacesTS} and fixed its weights during training.
We also adopted the learning rate warmup heuristic \cite{liu2019variance} and set the warmup step to 2000.
For dropout-based uncertainty estimation methods, we set the dropout-rate to 0.5.
The number of samples for Dropout are 50.
The number of audiences is 20 for our LEAD model.
$\zeta$ is set to 1.0. 
Our codes and datasets are available at \url{https://github.com/tshi04/DMSC_FEDA}.

\subsection{Prediction Performance}

\begin{table}[!t]
	\centering
	\caption{Averaged Accuracy (ACC) and MSE of different models on TripAdvisor-R (Trip-R), TripAdvisor-U (Trip-U), TripAdvisor-B (Trip-B), BeerAdvocate-R (Beer-R), and BeerAdvocate-B (Beer-B) testing sets. For MSE, smaller is better.
	$\dagger$ indicates that results are obtained from previous published papers and NA indicates that results are not available in those papers. we use bold font to highlight the best performance values and underline to highlight the second best values}
	\begin{tabular}{|l|c|c|c|c|c|c|c|c|}
		\hline
		\multirow{2}{*}{\bfseries Method}
		& \multicolumn{2}{c|}{\bf Trip-R}
		& \multicolumn{2}{c|}{\bf Trip-U}
		& \bf Trip-B
        & \multicolumn{2}{c|}{\bf Beer-R}
        & \bf Beer-B
		\\\cline{2-9}
		
		& \bf ACC & \bf MSE 
		& \bf ACC & \bf MSE 
		& \bf ACC
		& \bf ACC & \bf MSE 
		& \bf ACC
		\\\hline
		
		MAJOR
		& 29.12 & 2.115 & 39.73 & 1.222 & 62.42 
		& 26.29 & 4.252 & 67.26
		\\\hline

		GLVL
		& 38.94 & 1.795 & 48.04 & 0.879 & 78.15
		& 30.59 & 2.774 & 79.73
		\\\hline
		
		BOWL
		& 40.14 & 1.708 & 48.68 & 0.888 & 78.38
		& 31.02 & 2.715 & 79.14 
		\\\hline
		
		MCNN
		& 41.75 & 1.458 & 51.21 & 0.714 & 81.31 
		& 34.11 & 2.016 & 82.37 
		\\\hline
		
		MLSTM
		& 42.74 & 1.401 & 48.64 & 0.791 & 80.56
		& 34.48 & 2.167 & 82.07 
		\\\hline

		MATTN
		& 42.13 & 1.427 & 50.53 & 0.679 & 80.82
		& 35.78 & 1.962 & 84.86 
		\\\hline
		
		MBERT
		& 44.41 & 1.250
		& 54.50 & 0.617
		& 82.84 
		& 35.94 & 1.963 & 84.73
		\\\hline
		
		DMSCMC$\dagger$
		& 46.56 & \underline{1.083}
		& 55.49 & 0.583
		& \underline{83.34}
		& 38.06 & 1.755 & \underline{86.35}
		\\\hline
		
		HRAN$\dagger$ 
		& 47.43  & 1.169 
		& \underline{58.15} & \underline{0.528} 
		& NA
		& 39.11 & 1.700 & NA
		\\\hline
		
		AMN$\dagger$ 
		& \underline{48.66} & 1.109
		& NA & NA
		& NA
		& \underline{40.19} & \underline{1.686} & NA
		\\\hline
		
		FEDAR (Ours)
		& \bf 48.92 & \bf 1.072 & \bf 58.50 & \bf 0.522 & \bf 85.50 & \bf 40.62 & \bf 1.530 & \bf 87.40
		\\\hline
	\end{tabular}
	\label{tab:prediction_performance}
\end{table}

\begin{table}[!t]
	\centering
	\caption{Averaged accuracy (ACC) and MSE of different models on RateMDs-R (RMD-R) and RateMDs-B (RMD-B) testing sets. For MSE, smaller is better.}
		\begin{tabular}{|l|c|c|c|}
			\hline
            \multirow{2}{*}{\bfseries Method}
			& \multicolumn{2}{c|}{\bf RMD-R} 
			& {\bf RMD-B} 
			\\\cline{2-4}
			
			& \bf ACC & \bf MSE
			& \bf ACC
			\\\hline
			
			MAJOR
			& 31.42 & 3.393 & 57.18
			\\\hline
			
			GLVL
			& 43.11 & 1.882 & 76.93
			\\\hline
			
			BOWL
			& 44.78 & 1.704 & 78.68 \\\hline
			
			MCNN
			& 46.19	& 1.333 & 81.60
			\\\hline
			
			MLSTM
			& 48.37 & 1.148 & 82.40
			\\\hline
			
			MATTN
			& 49.08 & 1.157 & 82.66
			\\\hline
			
			MBERT
			& 48.65 & 1.160 & 83.39
			\\\hline
			
			FEDAR (Ours)
			& \bf 55.57 & \bf 0.794 & \bf 88.63
			\\\hline
	\end{tabular}
	\label{tab:expRatemds}
\end{table}

For research question \textbf{RQ1}, we use accuracy (ACC) and mean squared error (MSE) as our evaluation metrics to measure the prediction performance of different models.
All results are shown in Tables~\ref{tab:prediction_performance} and \ref{tab:expRatemds}, where we use bold font to highlight the best performance values and underline to highlight the second best values.

For the DMSC problem, it has been demonstrated that deep neural network (DNN) based models perform much better than conventional machine learning methods that rely on $n$-gram or embedding features \cite{yin2017document,li2018document}.
In our experiments, we have also demonstrated this by comparing different DNN models with MAJOR, GLVL and BOWL.
Compared to simple DNN classification models, multi-task learning DNN models (MDNN) can achieve better results with fewer parameters and training time \cite{yin2017document}.
Therefore, we focused on comparing the performance of our model with different MDNN models.
From Table~\ref{tab:prediction_performance}, DMSCMC achieves better results on all five datasets compared with baselines MCNN, MLSTM, MBERT, and MATTN.
HRAN and AMN leverage the power of overall rating and get significantly better results than other compared methods.
From both tables, we observed our FEDAR model achieves the best performance on all seven datasets. 
These results demonstrate the effectiveness of our methods.

\subsection{Uncertainty Performance}

\begin{table}[!t]
    \centering
    \caption{
    Performance of various uncertainty methods on different datasets.}
    \begin{tabular}{|l|c|c|c|c|c|}
        \multicolumn{6}{c}{\bf TripAdvisor-R}\\
        \hline
        \bf Method & \bf top-5\% & \bf top-10\% & \bf top-15\% & \bf top-20\% & \bf top-25\% \\
        \hline
        Max-Margin &
        35.40 & 36.00 & 37.47 & 39.15 & 40.68
        \\\hline
        PL-Variance &
        40.20 & 42.40 & 43.00 & 44.25 & 44.84
        \\\hline
        Dropout & 
        53.80 & 53.50 & 53.33 & 53.35 & 53.60
        \\\hline
        LEAD  & 
        \bf 65.40 & \bf 62.60 & \bf 60.93 & \bf 60.85 & \bf 60.20
        \\\hline
        
        \multicolumn{6}{c}{\bf BeerAdvocate-R}\\\hline
        \bf Method & \bf top-5\% & \bf top-10\% & \bf top-15\% & \bf top-20\% & \bf top-25\% \\
        \hline
        Max-Margin &
        38.80 & 43.80 & 46.53 & 48.30 & 49.44
        \\\hline
        PL-Variance &
        44.00 & 47.00 & 48.33 & 49.65 & 50.68
        \\\hline
        Dropout & 
        57.00 & 57.90 & 58.60 & 58.70 & 59.28
        \\\hline
        LEAD & 
        \bf 71.60 & \bf 69.50 & \bf 67.93 & \bf 67.55 & \bf 67.20
        \\\hline
        
        \multicolumn{6}{c}{\bf RateMDs-R}\\\hline
        \bf Method & \bf top-5\% & \bf top-10\% & \bf top-15\% & \bf top-20\% & \bf top-25\% \\
        \hline
        Max-Margin & 20.20 & 23.80 & 26.80 & 28.15 & 29.40
        \\\hline
        
        PL-Variance & 28.60 & 29.70 & 30.53 & 30.85 & 31.60
        \\\hline
        
        Dropout & 51.00 & 50.70 & 50.60 & 49.60 & 48.88
        \\\hline
        
        LEAD & \bf 66.00 & \bf 62.70 & \bf 60.40 & \bf 59.05 & \bf 58.32
        \\\hline
    \end{tabular}
    \label{tab:uncertainty_performance}
\end{table}

Uncertainty estimation can help users identify reviews for which the models are not confident of their predictions.
More intuitively, prediction models are prone to mistakes on the reviews that they are uncertain about.
In Table \ref{tab:uncertainty_performance}, we first selected the most uncertain predictions (denoted by \textbf{top-n\%}) based uncertainty scores from the testing sets of TripAdvisor-R, BeerAdvocate-R and RateMDs-R datasets.
Then, we evaluated the uncertainty performance by comparing the mis-classification rate (i.e., error rate) of our FEDAR model for the selected reviews.
The more incorrect predictions that can be captured, the better the uncertainty method will be.
From these results, we can observe that Dropout method achieves significantly better results than Max-Margin and PL-Variance.
Our LEAD method outperforms all these baseline methods on three datasets, which shows our method is superior in identifying less confident predictions and answers research question \textbf{RQ2}.

\subsection{Ablation Study of FEDAR}

\begin{table}[!t]
	\centering
	\caption{Ablation study results. Different models are evaluated by Averaged Accuracy (ACC) and MSE metrics on five public DMSC testing sets. For MSE, smaller is better. \textbf{FE}, \textbf{DA} and \textbf{OR} represent Feature Enrichment, Deliberated self-Attention, Overall Rating, respectively.
	}
	\begin{tabular}{|l|c|c|c|c|c|c|c|c|}
		\hline
		\multirow{2}{*}{\bfseries Method}
		& \multicolumn{2}{c|}{\bf Trip-R}
		& \multicolumn{2}{c|}{\bf Trip-U}
		& \bf Trip-B
        & \multicolumn{2}{c|}{\bf Beer-R}
        & \bf Beer-B
		\\\cline{2-9}
		
		& \bf ACC & \bf MSE 
		& \bf ACC & \bf MSE 
		& \bf ACC
		& \bf ACC & \bf MSE 
		& \bf ACC
		\\\hline
		
		FEDAR
		& \bf 48.92 & \bf 1.072 & \bf 58.50 & \bf 0.522 & \bf 85.50 & \bf 40.62 & \bf 1.530 & \bf 87.40
		\\\hline
		
		w/o OR
		& 46.72 & 1.178 & 55.82 & 0.574 & {84.23} 
		& 39.66 & {1.617} & {86.52}
		\\\hline
		
		w/o OR, DA
		& 45.70 & 1.224 & 55.39 & 0.584 & 83.43
		& 38.85 & 1.633 & 85.99
		\\\hline
		
		w/o OR, FE
		& 44.50 & 1.300 & 53.41 & 0.632 & 82.39
		& 38.92 & 1.714 & 84.99
		\\\hline
		
		w/o OR, DA, FE 
		& 42.13 & 1.427 & 50.53 & 0.679 & 80.82
		& 35.78 & 1.962 & 84.86
		\\\hline
		
	\end{tabular}
	\label{tab:ablation1}
\end{table}

\begin{table}[!t]
	\centering
	\caption{Ablation study results. Different models are evaluated by Averaged Accuracy (ACC) and MSE metrics on RateMDs-R (RMD-R) and RateMDs-B (RMD-B) testing sets.
	}
	\begin{tabular}{|l|c|c|c|}
		\hline
		\multirow{2}{*}{\bfseries Method}
        & \multicolumn{2}{c|}{\bf RMD-R}
        & \bf RMD-B
		\\\cline{2-4}
		
		& \bf ACC & \bf MSE 
		& \bf ACC
		\\\hline
		
		FEDAR
		& \bf 55.82 & \bf 0.786 & \bf 88.63
		\\\hline
		
		w/o OR
		& 49.80 & 1.106 & 83.89
		\\\hline
		
		w/o OR, DA
		& 49.68 & 1.108 &  83.62 
		\\\hline
		
		w/o OR, FE
		& 49.28 & 1.123 & 83.47
		\\\hline
		
		w/o OR, DA, FE
		& 49.08 & 1.157 & 82.66
		\\\hline
		
	\end{tabular}
	\label{tab:ablation2}
\end{table}

\begin{figure}[!tp]
    \centering
    \begin{subfigure}[b]{0.4\linewidth}
    \includegraphics[width=\linewidth]{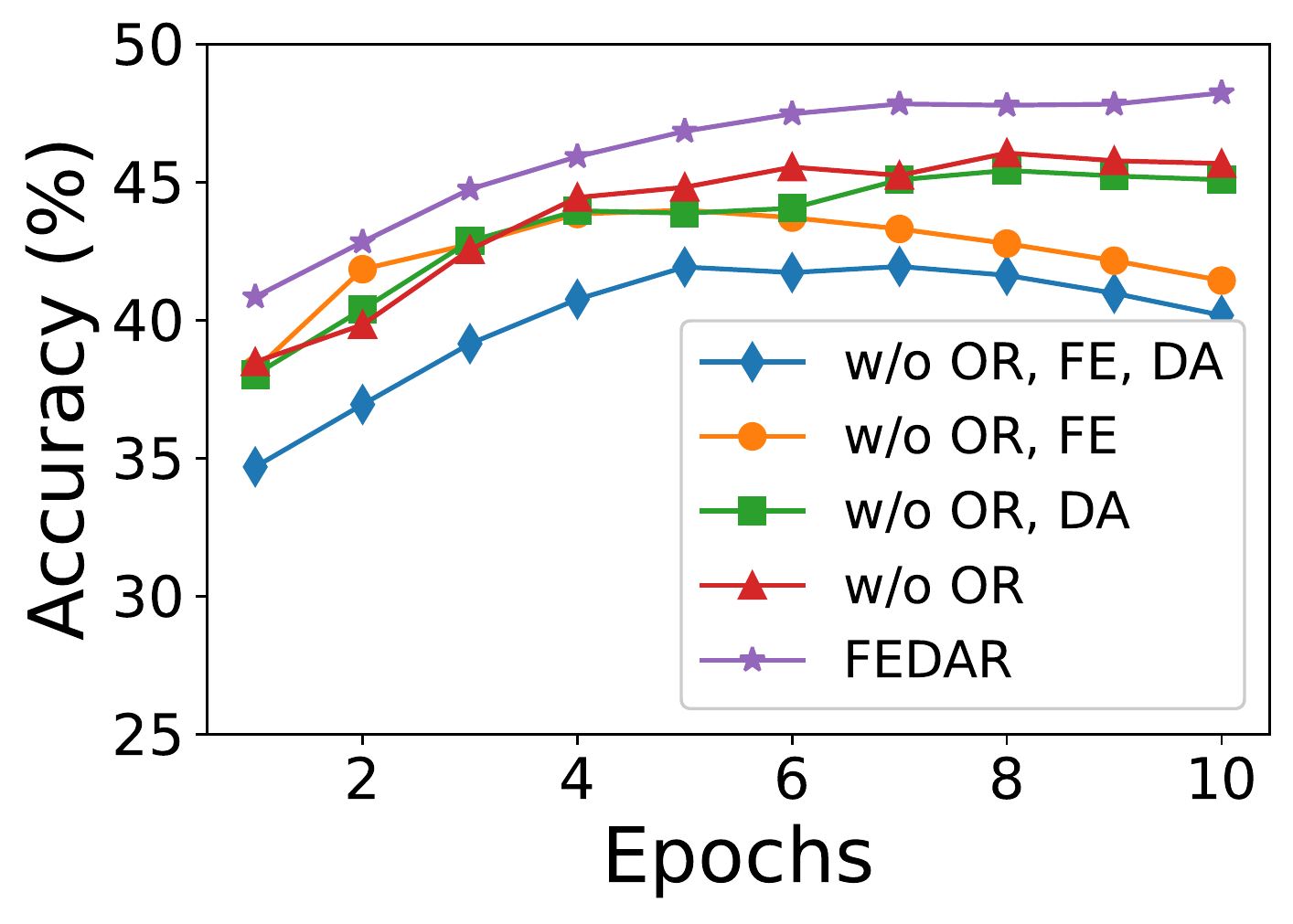}
    \caption{Accuracy vs. Epochs}
    \end{subfigure}
    \begin{subfigure}[b]{0.4\linewidth}
    \includegraphics[width=\linewidth]{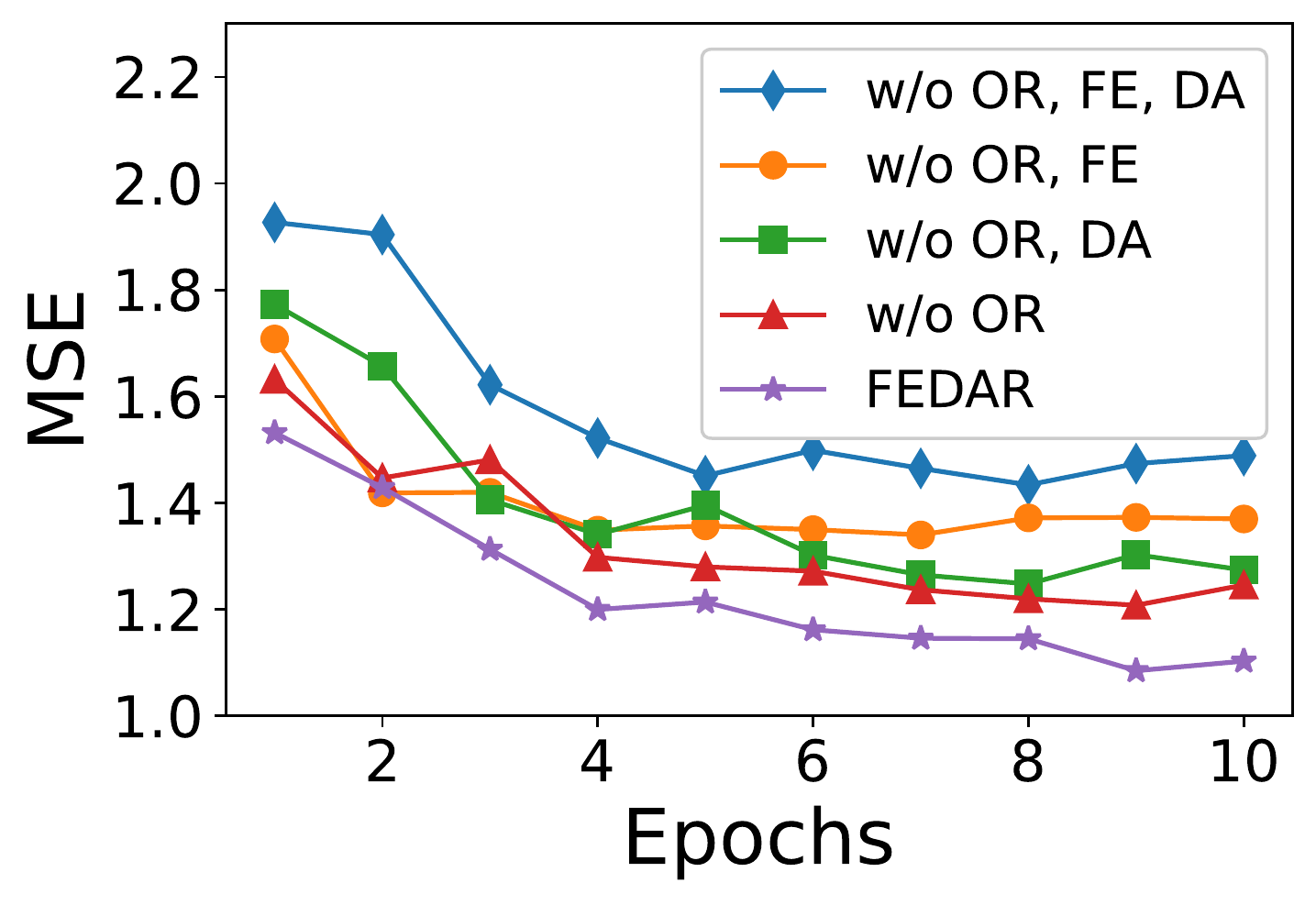}
    \caption{MSE vs. Epochs}
    \end{subfigure}
    
    \caption{This figure shows (a) Averaged Accuracy and (b) MSE for FEDAR and its variants on the TripAdvisor-R dataset during the training process.}
    \label{fig:ablation}
\end{figure}

For research question \textbf{RQ3}, we attribute the performance improvement of our FEDAR model to: 1) Better review encoder, including a highway word embedding layer and a feature enriched encoder.
2) Deliberate self-attention mechanism.
3) Overall rating.
Therefore, we systematically conducted ablation studies to demonstrate the effectiveness of these components, and provided the results in Table~\ref{tab:ablation1}, Table~\ref{tab:ablation2} and Fig.~\ref{fig:ablation}.

We first observe that FEDAR significantly outperforms model-OR (FEDAR w/o OR), which indicates that overall rating can help the model make better predictions.
Secondly, we compare model-OR with model-ORFE (FEDAR w/o OR, FE), which is equipped with a regular word embedding layer and a multi-layer Bi-LSTM encoder.
Obviously, model-OR obtained better results than model-ORFE.
Similarly, we also compare model-ORDA (FEDAR w/o OR, DA) with model-BASE (FEDAR w/o OR, DA, FE), since model-ORDA adopts the same self-attention mechanism as model-BASE.
It can be observed that model-ORDA performs significantly better than model-BASE on all the datasets.
This experiment shows that we can improve the performance by using highway word embedding layer and feature enrichment technique.
Furthermore, we compared model-OR
with model-ORDA, which does not have a deliberate self-attention layer.
It can been seen that model-OR outperforms model-ORDA in all the experiments.
In addition, we have also compared the results of model-ORFE and model-BASE, which are equipped with a deliberate self-attention layer and a regular self-attention layer.
We observed that model-ORFE has a better performance compared to model-BASE.
This experiment indicates the effectiveness of deliberate self-attention mechanism.
In Fig.~\ref{fig:ablation}, we show the accuracy and MSE of different models during training in order to demonstrate that FEDAR can get consistently higher accuracy and lower MSE after training for several epochs than its basic variants.

\subsection{Attention Visualization}

The attention mechanism enables a model to selectively focus on important parts of the reviews, and hence, visualization of the attention weights can help in interpreting our model and analyze the experimental results \cite{yin2017document,xu2015show}. 
To answer research question \textbf{RQ4}, we need to investigate whether our model attends to relevant keywords when it is making aspect-specific rating predictions for the DMSC problem.

\begin{figure}[!tp]
    \centering
    \begin{subfigure}[b]{0.5\linewidth}
    \includegraphics[width=\linewidth]{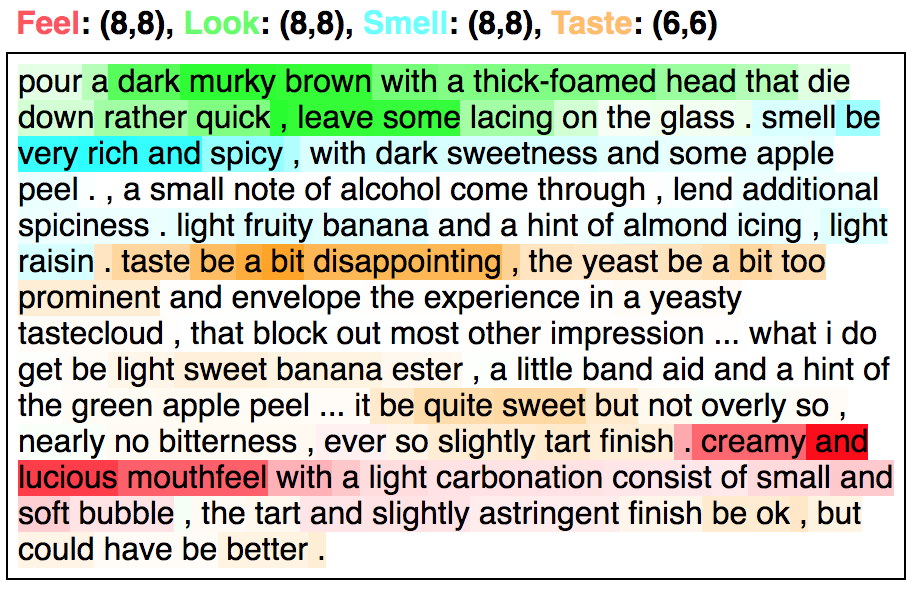}
    \caption{BeerAdvocate-R}
    \end{subfigure}
    \\
    \begin{subfigure}[b]{0.5\linewidth}
    \includegraphics[width=\linewidth]{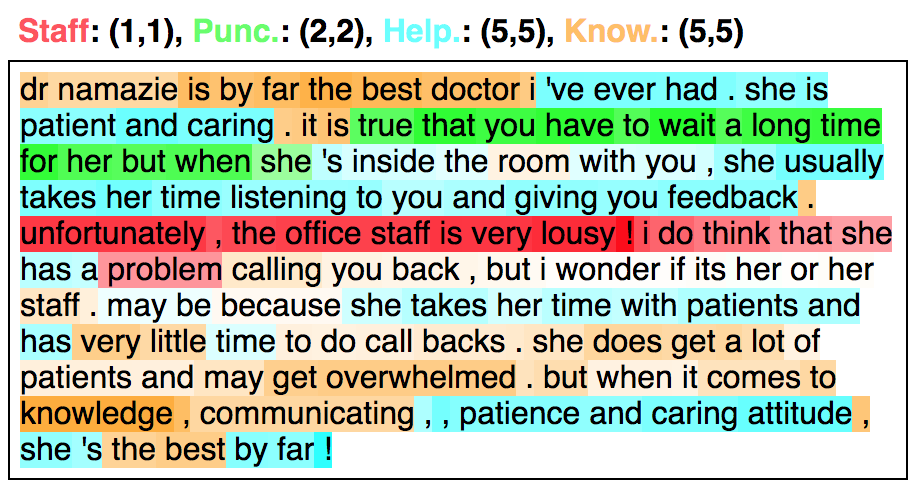}
    \caption{RateMDs-R}
    \end{subfigure}
    
    \caption{Visualization of attention weights. In parentheses, the first and second numbers represent ground-truth and predicted ratings, respectively.
	Different aspects are labeled with different colors. The figure is best viewed in color.}
    \label{fig:fine_example}
\end{figure}

In Fig.~\ref{fig:fine_example} (a), we show a review example from the BeerAdvocate-R testing set, for which our model has successfully predicted all aspect-specific ratings.
In this Figure, we highlighted the review with deliberate attention weights.
The review contains keywords of all four aspects, thus, we only need to verify whether our model can successfully detect those aspect-specific keywords.
We observed that deliberate self-attention attend to ``{\it creamy and luscious mouthfeel}" for {\bf feel}.
For the {\bf look} aspect, it captures ``{\it dark murky brown with a ..., leave some lacing on the glass}'', which is quite relevant to the appearance of the beer.
Our model also successfully detects ``{\it very rich and spicy}" for {\bf smell}.
For {\bf taste}, it attends to ``{\it taste is a bit disappointing, ... too prominent}", which yields a slightly lower rating.
Similarly, we show an example from the RateMDs-R testing set in Fig~.\ref{fig:fine_example} (b).
Our model detects ``\textit{unfortunately, the office staff is very lousy! I do think ...}'' for \textbf{staff}, which expresses negative opinion to the office staff.
For \textbf{punctuality}, it captures ``\textit{true that you have to wait a long time for her}'', which is also negative.
Finally, it attends to ``\textit{is by far the best doctor, she does get a lot of patient and may get overwhelmed. but when it comes to knowledge, communicating, the best}'' for the \textbf{knowledge}, and ``\textit{she is patient and caring, patience and caring attitude}'' for the \textbf{helpfulness} of the doctor.
Both aspects have positive sentiment. 
Therefore, these two examples show good interpretability of our model.

\subsection{Aspect and Opinion Keywords}

\begin{figure}[!t]
    \centering
    
    \resizebox{1\linewidth}{!}{
        \begin{subfigure}[b]{0.24\linewidth}
        \includegraphics[width=\linewidth]{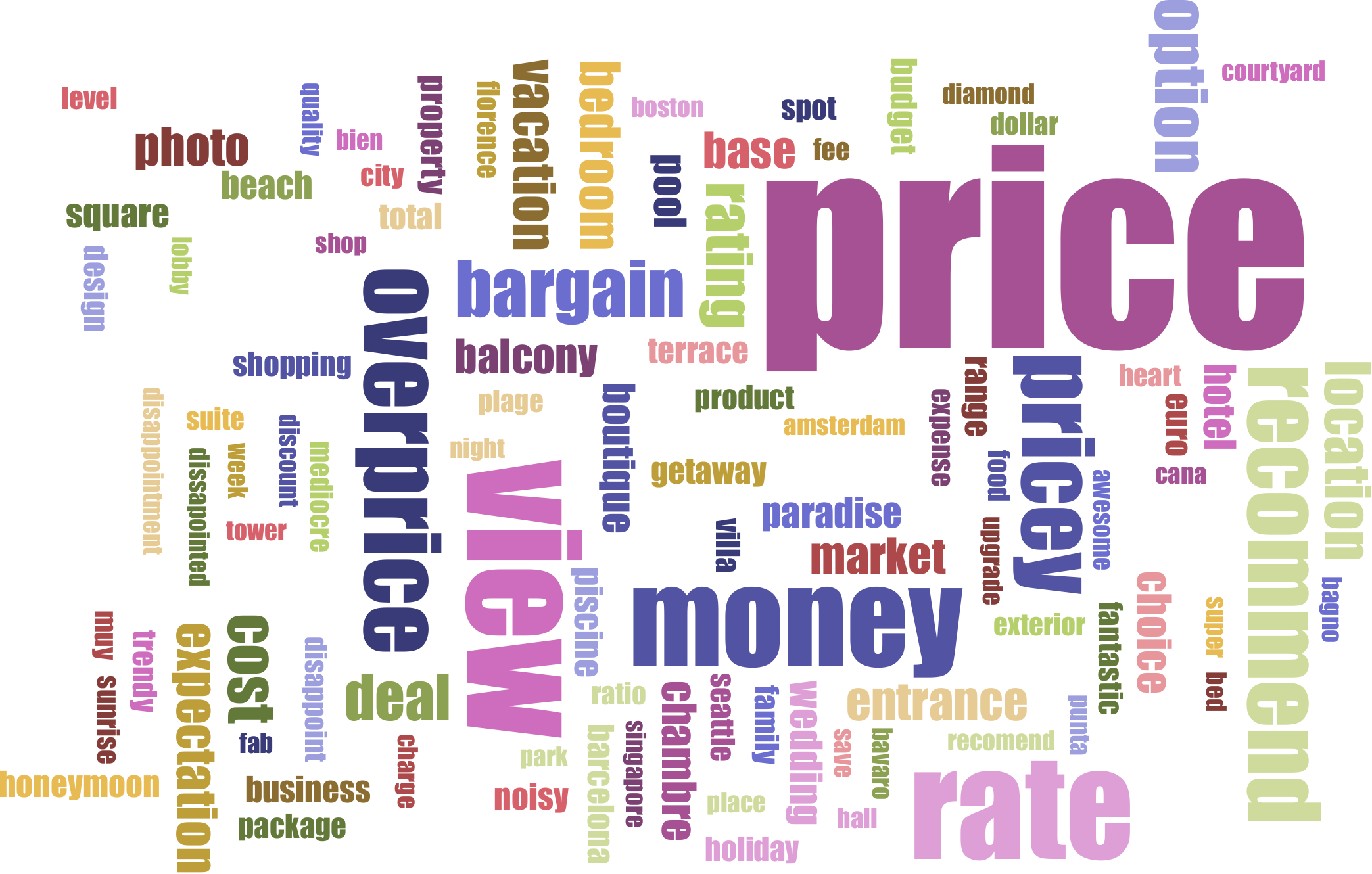}
        \caption{Value}
        \end{subfigure}
        \begin{subfigure}[b]{0.24\linewidth}
        \includegraphics[width=\linewidth]{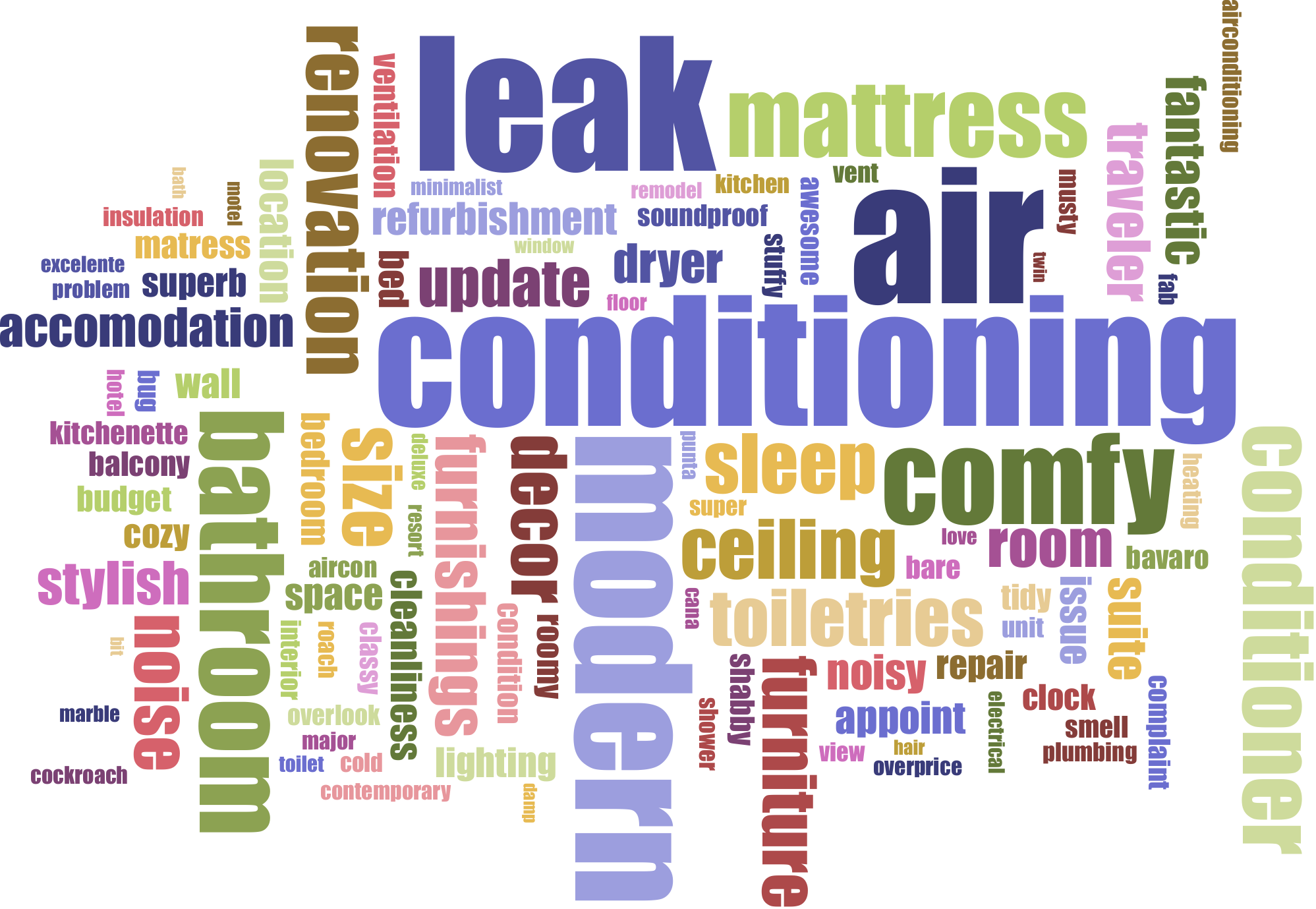}
        \caption{Room}
        \end{subfigure}
        \begin{subfigure}[b]{0.24\linewidth}
        \includegraphics[width=\linewidth]{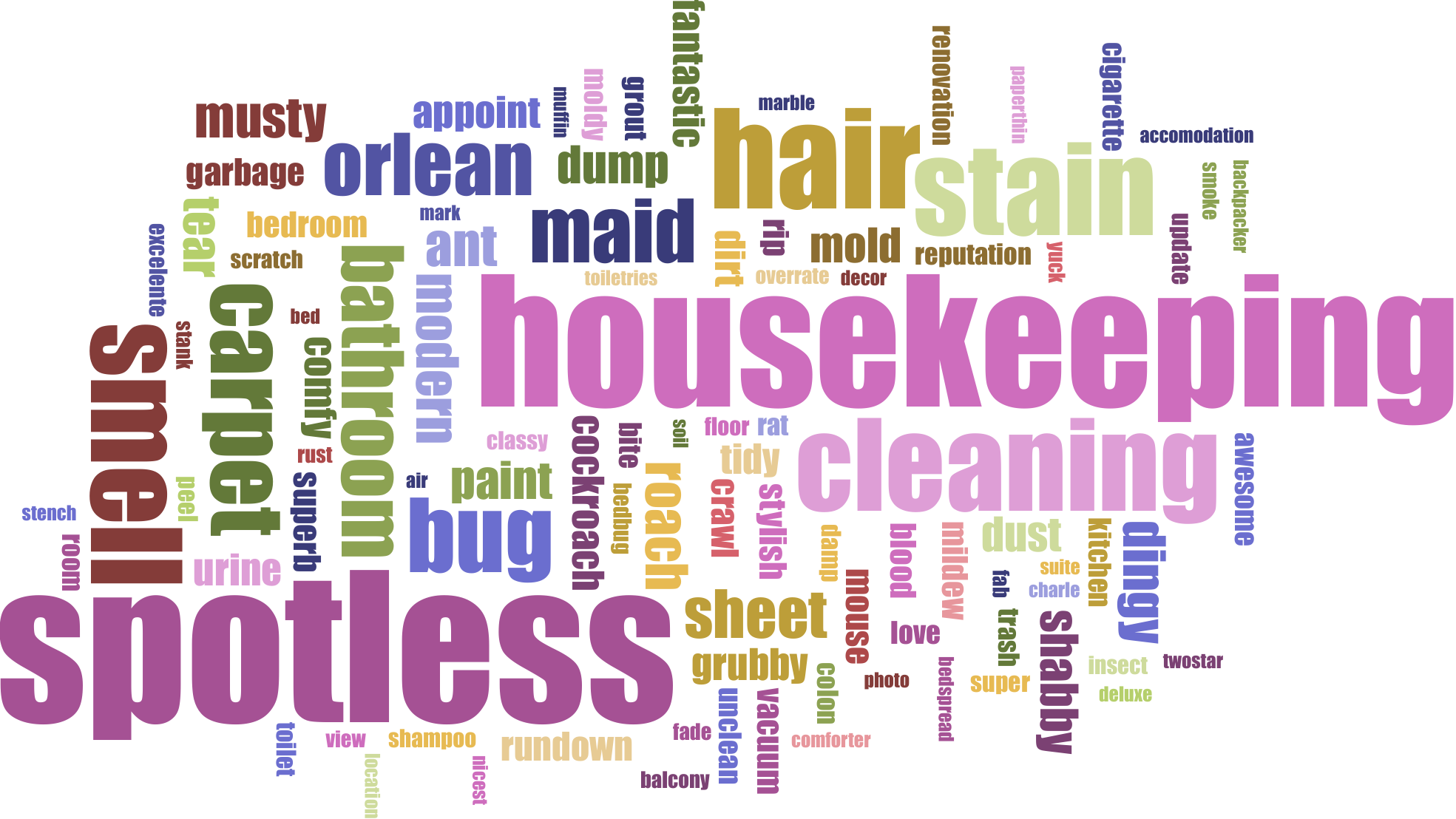}
        \caption{Cleanliness}
        \end{subfigure}
        \begin{subfigure}[b]{0.24\linewidth}
        \includegraphics[width=\linewidth]{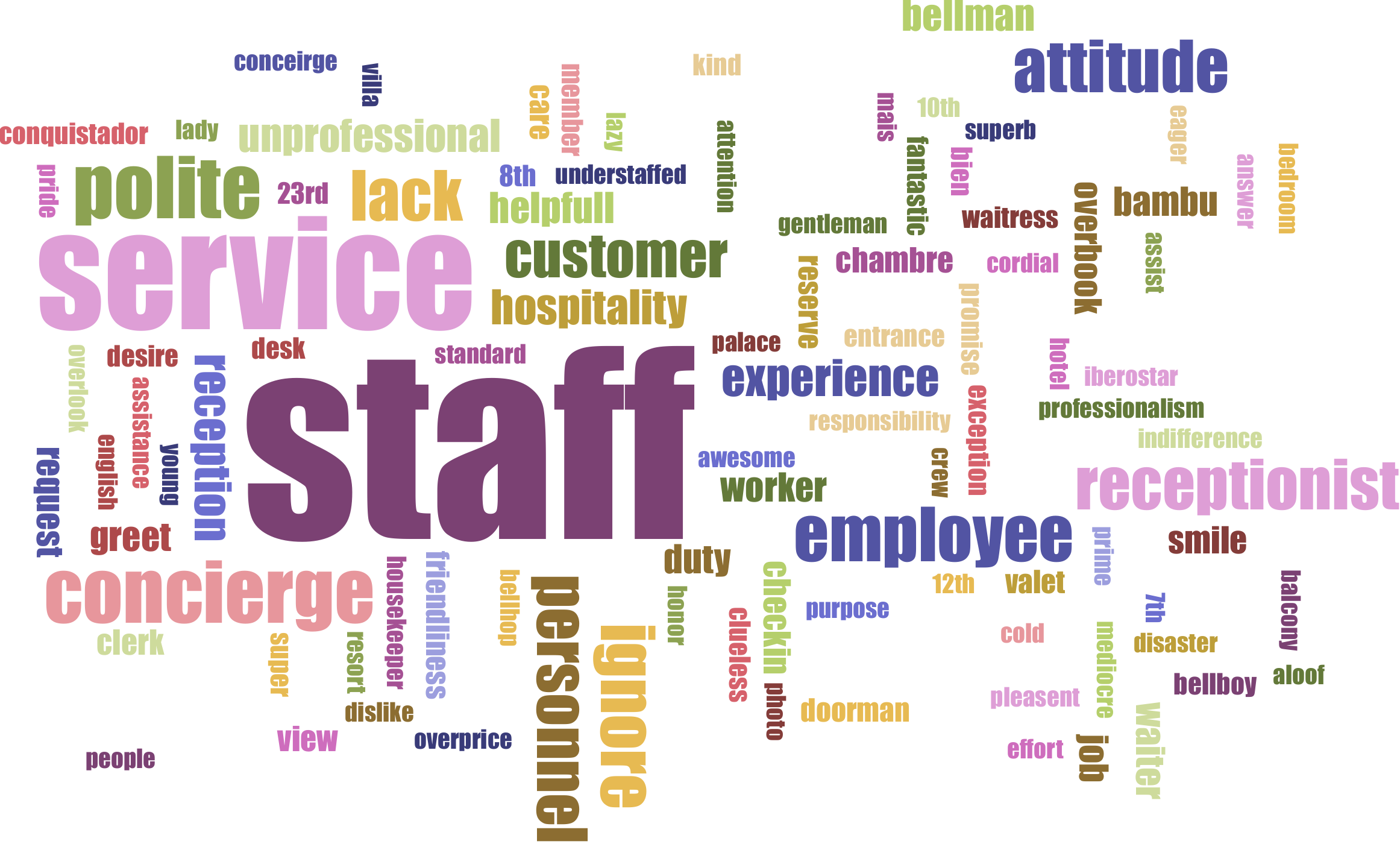}
        \caption{Service}
        \end{subfigure}}
    
    \resizebox{1\linewidth}{!}{
        \begin{subfigure}[b]{0.24\linewidth}
        \includegraphics[width=\linewidth]{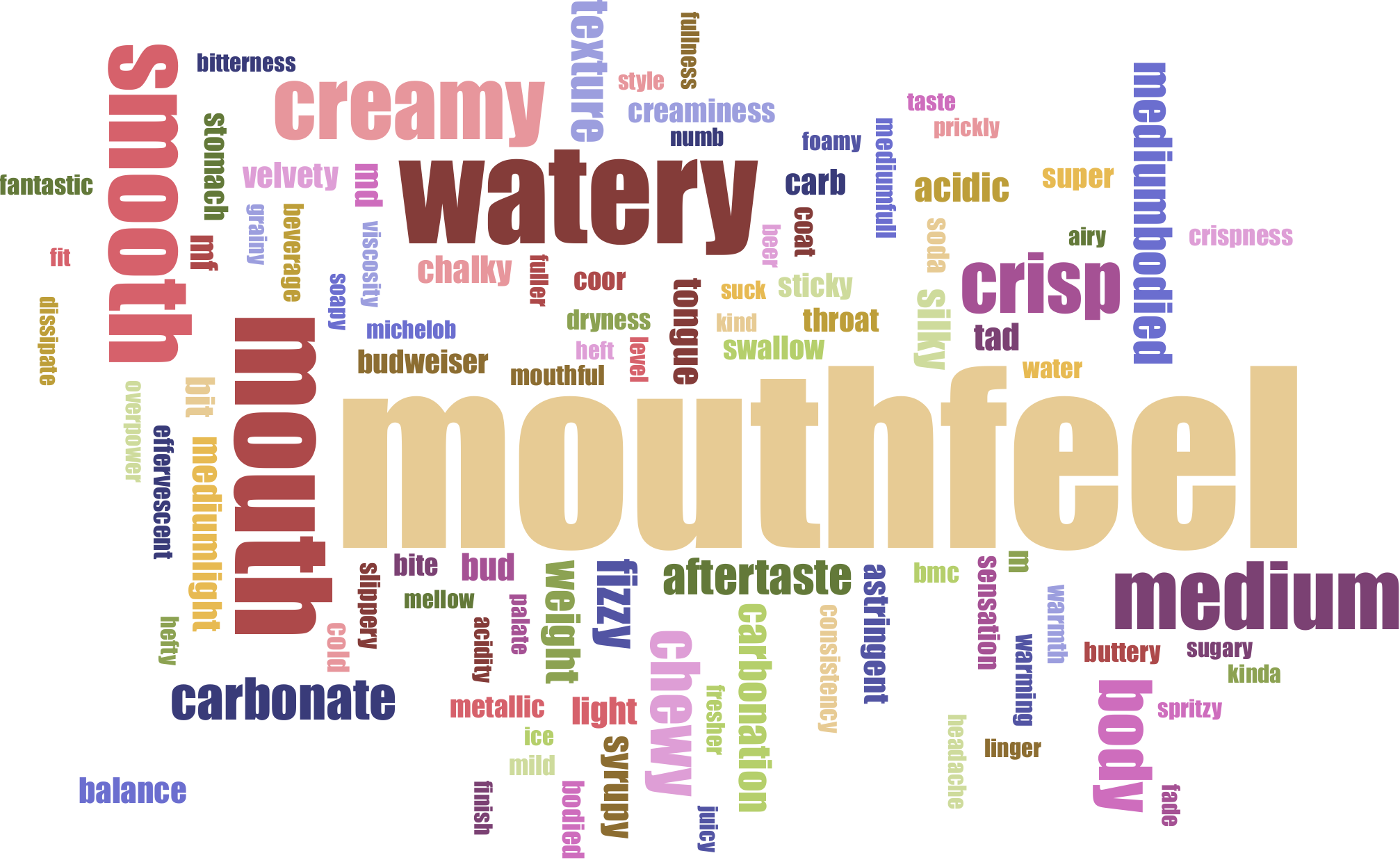}
        \caption{Feel}
        \end{subfigure}
        \begin{subfigure}[b]{0.24\linewidth}
        \includegraphics[width=\linewidth]{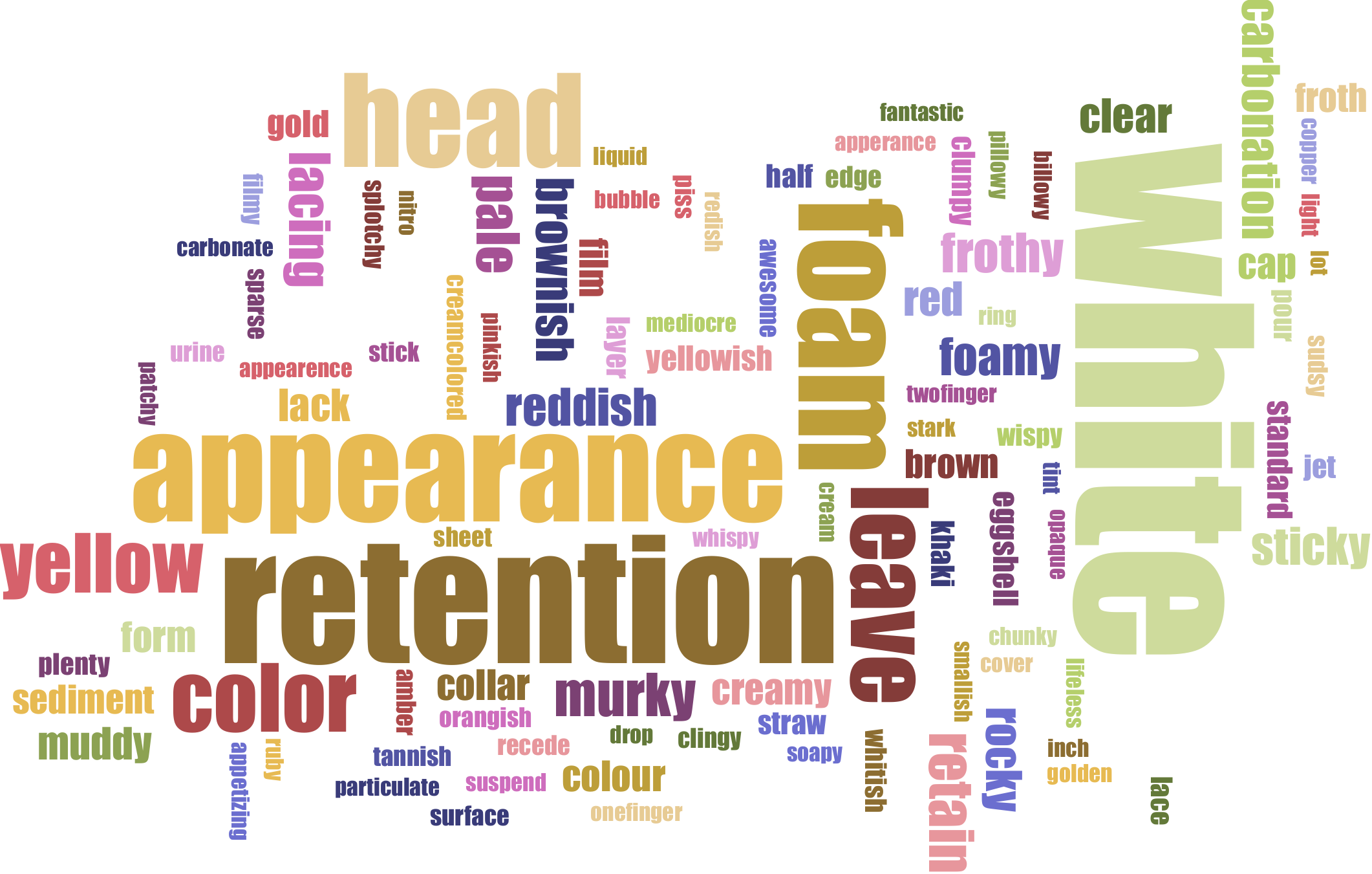}
        \caption{Look}
        \end{subfigure}
        \begin{subfigure}[b]{0.24\linewidth}
        \includegraphics[width=\linewidth]{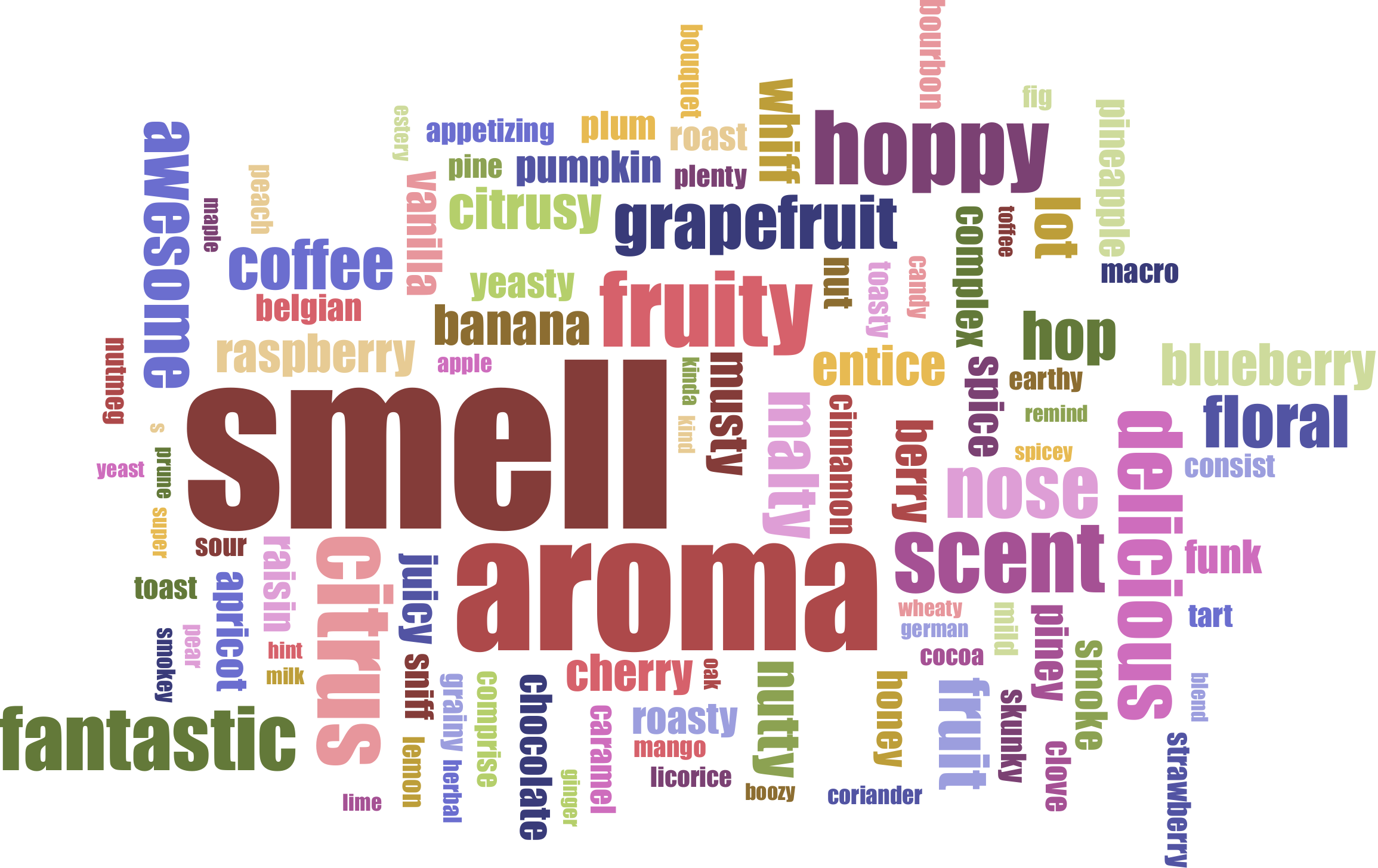}
        \caption{Smell}
        \end{subfigure}
        \begin{subfigure}[b]{0.24\linewidth}
        \includegraphics[width=\linewidth]{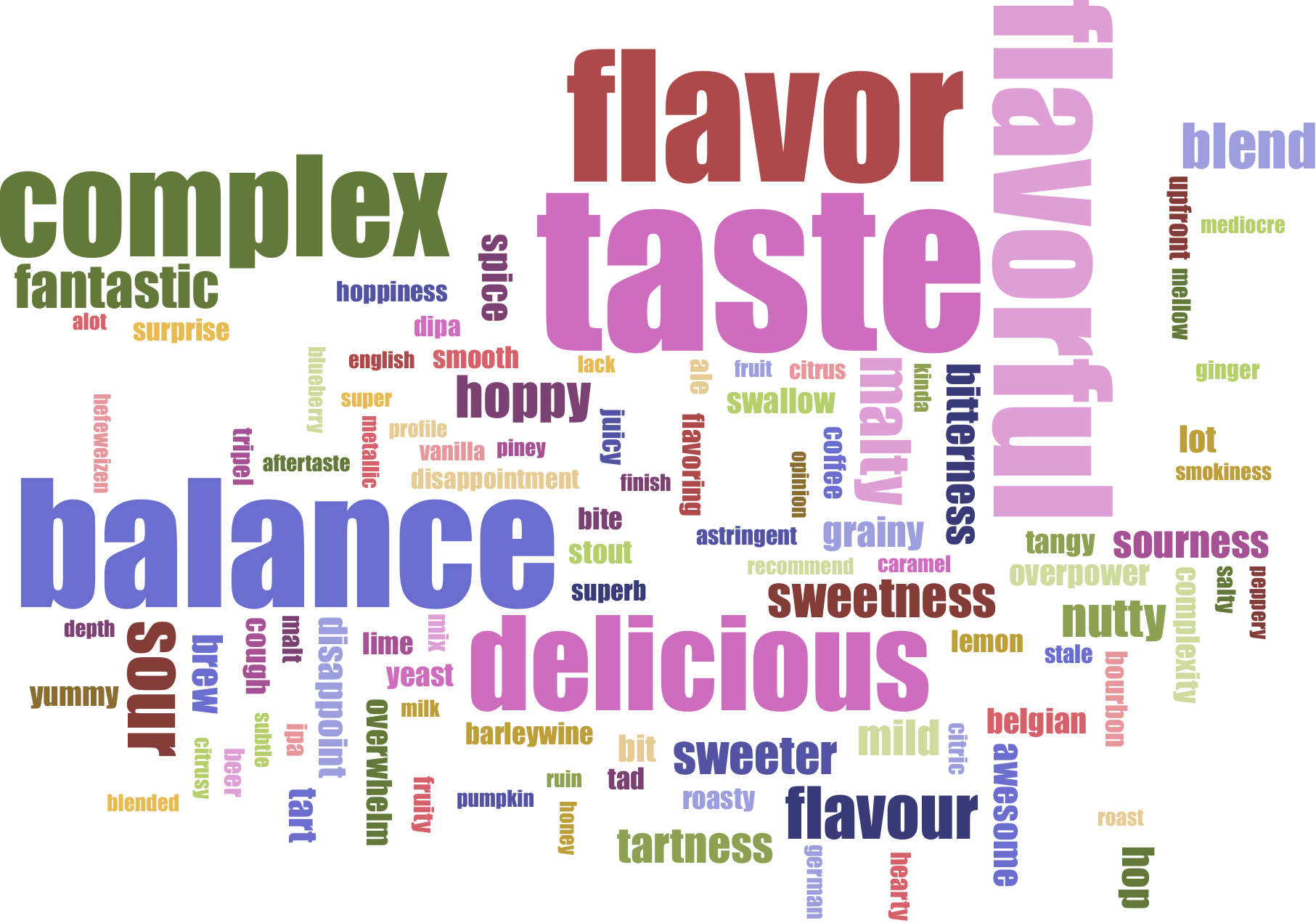}
        \caption{Taste}
        \end{subfigure}}
        
        \resizebox{1\linewidth}{!}{
        \begin{subfigure}[b]{0.24\linewidth}
        \includegraphics[width=\linewidth]{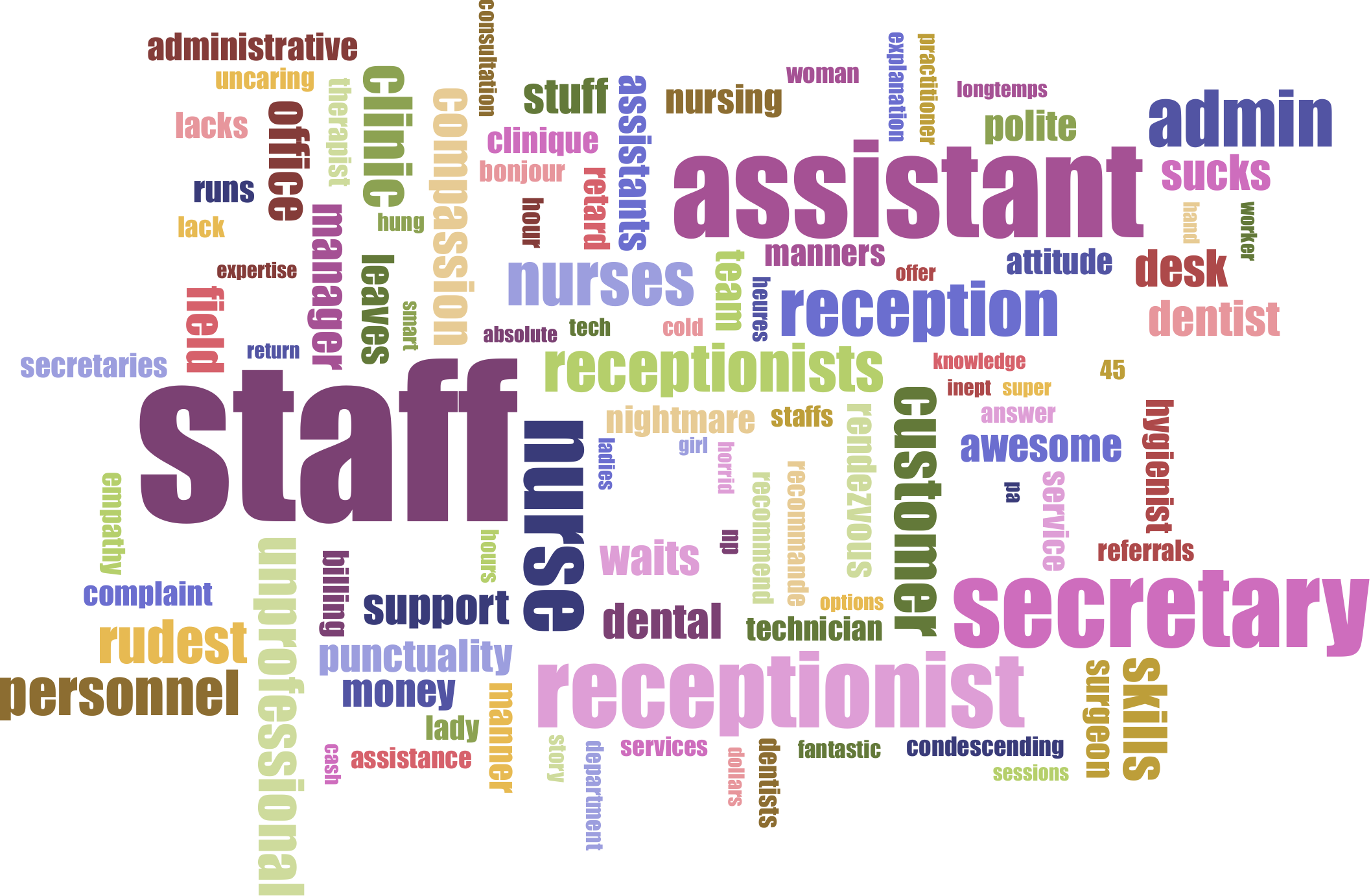}
        \caption{Staff}
        \end{subfigure}
        \begin{subfigure}[b]{0.24\linewidth}
        \includegraphics[width=\linewidth]{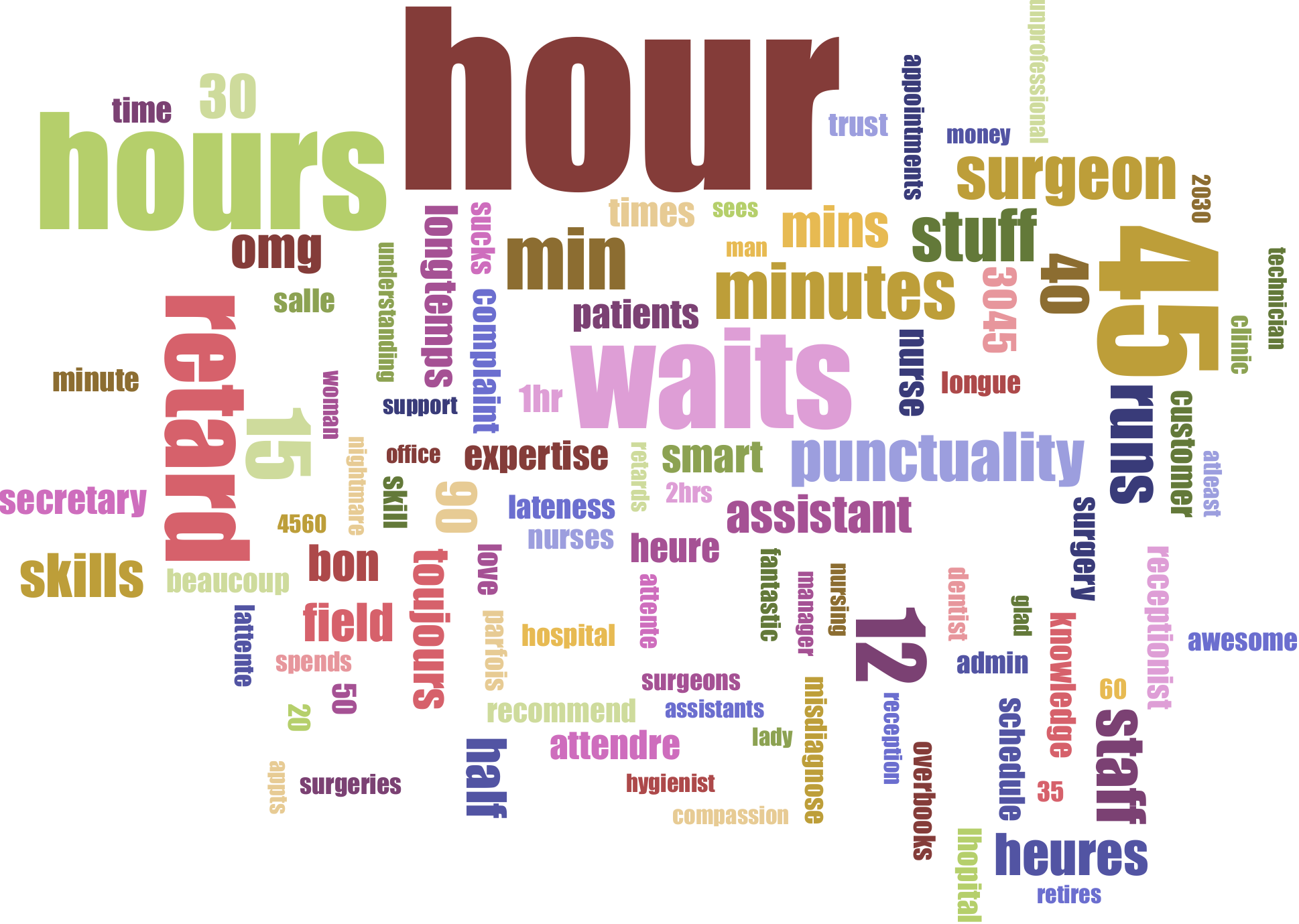}
        \caption{Punctuality}
        \end{subfigure}
        \begin{subfigure}[b]{0.24\linewidth}
        \includegraphics[width=\linewidth]{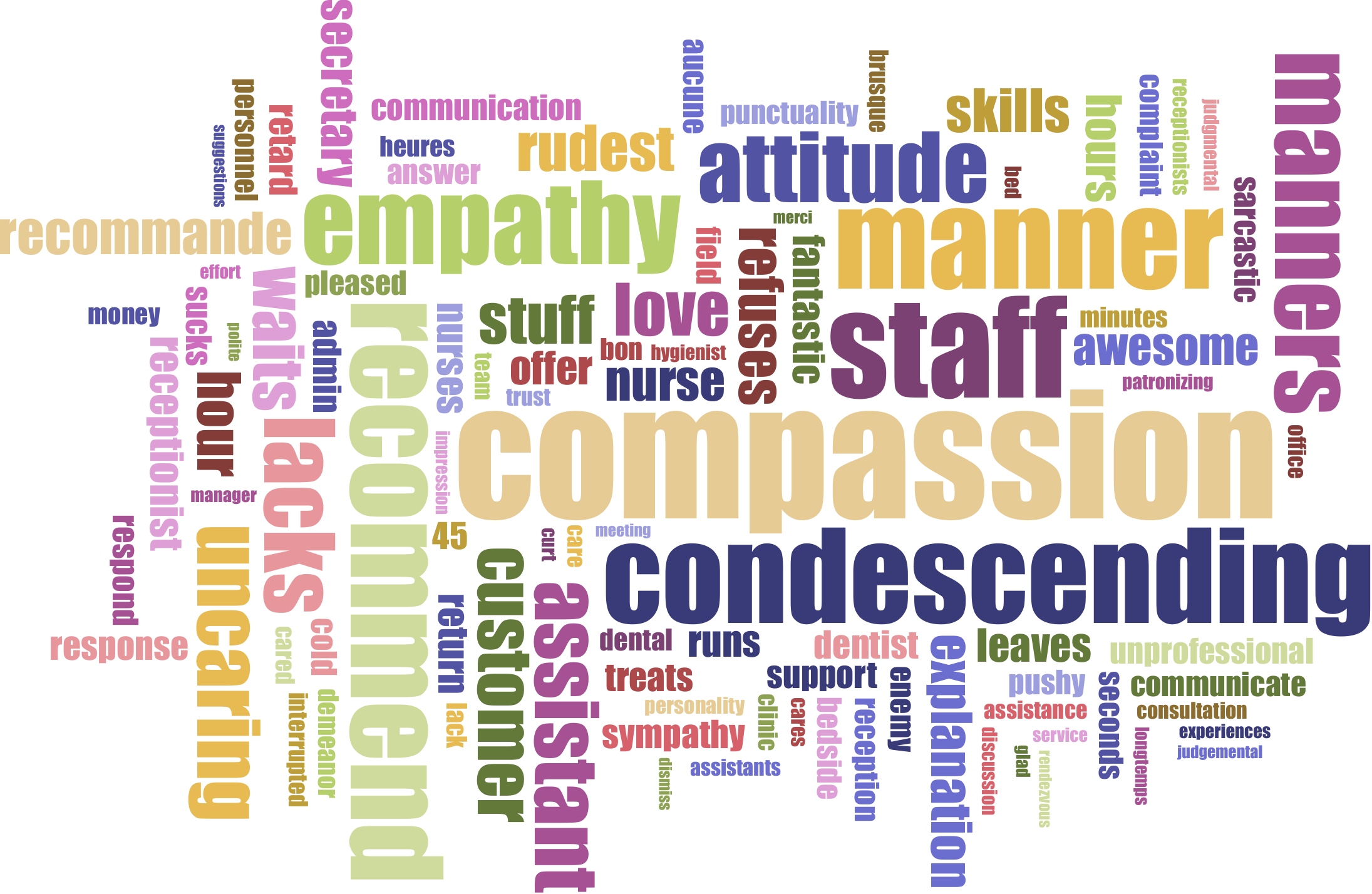}
        \caption{Helpfulness}
        \end{subfigure}
        \begin{subfigure}[b]{0.24\linewidth}
        \includegraphics[width=\linewidth]{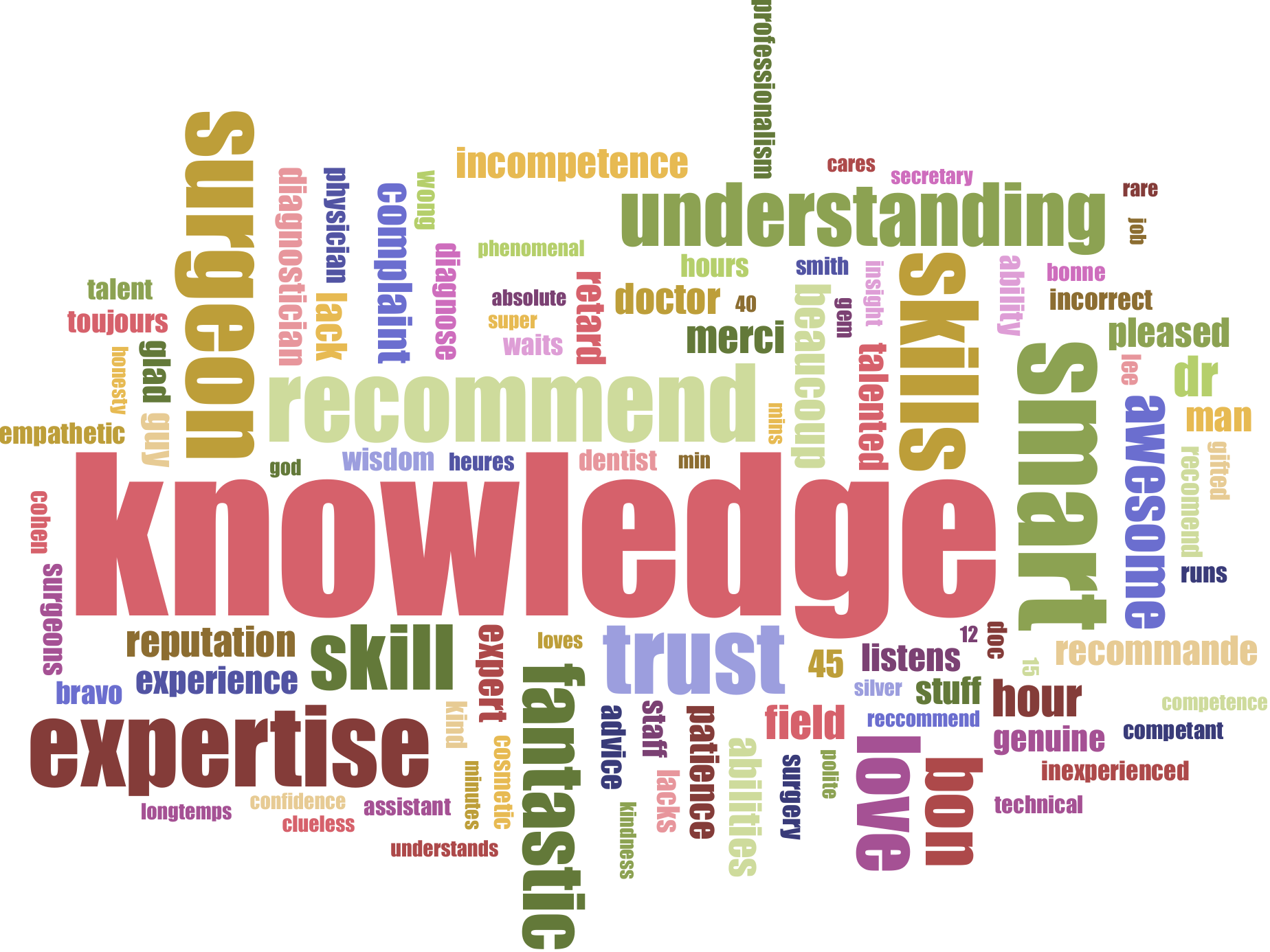}
        \caption{Knowledge}
        \end{subfigure}}
    
    \caption{Word-cloud visualization of aspect keywords for TripAdvisor-B (Top row), BeerAdvocate-B (Middle row) and RateMDs-B datasets (Bottom row).}
    \label{fig:word_cloud_aspect_keywords}
\end{figure}

In Fig.~\ref{fig:word_cloud_aspect_keywords}, we first show aspect keywords detected by our AKR method for TripAdivsor-B, BeerAdvocate-B, and RateMDs-B corpus.
From Fig.~\ref{fig:word_cloud_aspect_keywords} (Top row), we observe that \textbf{value} related keywords include ``\textit{price, money, rate, overprice}''.
Keywords related to a \textbf{room} are ``\textit{air conditioning, comfy, leak, mattress, bathroom, modern, ceiling}'' and others.
For \textbf{cleanliness}, people are interested in ``\textit{housekeeping, spotless, cleaning, hair, stain, smell}'' and so on. \textbf{Service} is related with ``\textit{staff, service, employee, receptionist, personnel}''.
From Fig.~\ref{fig:word_cloud_aspect_keywords} (Middle row), we observe that {\bf feel} is usually related with keywords, like ``{\it mouthfeel, mouth, smooth, watery}'', which describe feel of beers in mouth.
{\bf Look} is the appearance of beers, thus, the model captures ``{\it appearance, retention, white, head, foam, color}'' and others.
{\bf Smell} related aspect keywords include ``\textit{smell, aroma, scent, fruity}'' and more.
Finally, representative keywords for {\bf taste} are ``{\it taste, balance, complex, flavor}'' and so on.
From Fig.~\ref{fig:word_cloud_aspect_keywords} (Bottom row), we observe that \textbf{staff} related keywords are ``\textit{staff, assistant, secretary, receptionist}'' and so on.
For \textbf{punctuality}, people usually concern ``\textit{waits, hour, hours, retard}''.
The \textbf{helpfulness} of a doctor is related to ``\textit{compassion, manner, empathy, attitude, condescending}'' and so on.
Finally, \textbf{knowledge} related keywords are ``\textit{knowledge, expertise, surgeon, skill}'' and others.

\begin{figure}[!t]
    \centering
    
    \resizebox{1\linewidth}{!}{
        \begin{subfigure}[b]{0.24\linewidth}
        \includegraphics[width=\linewidth]{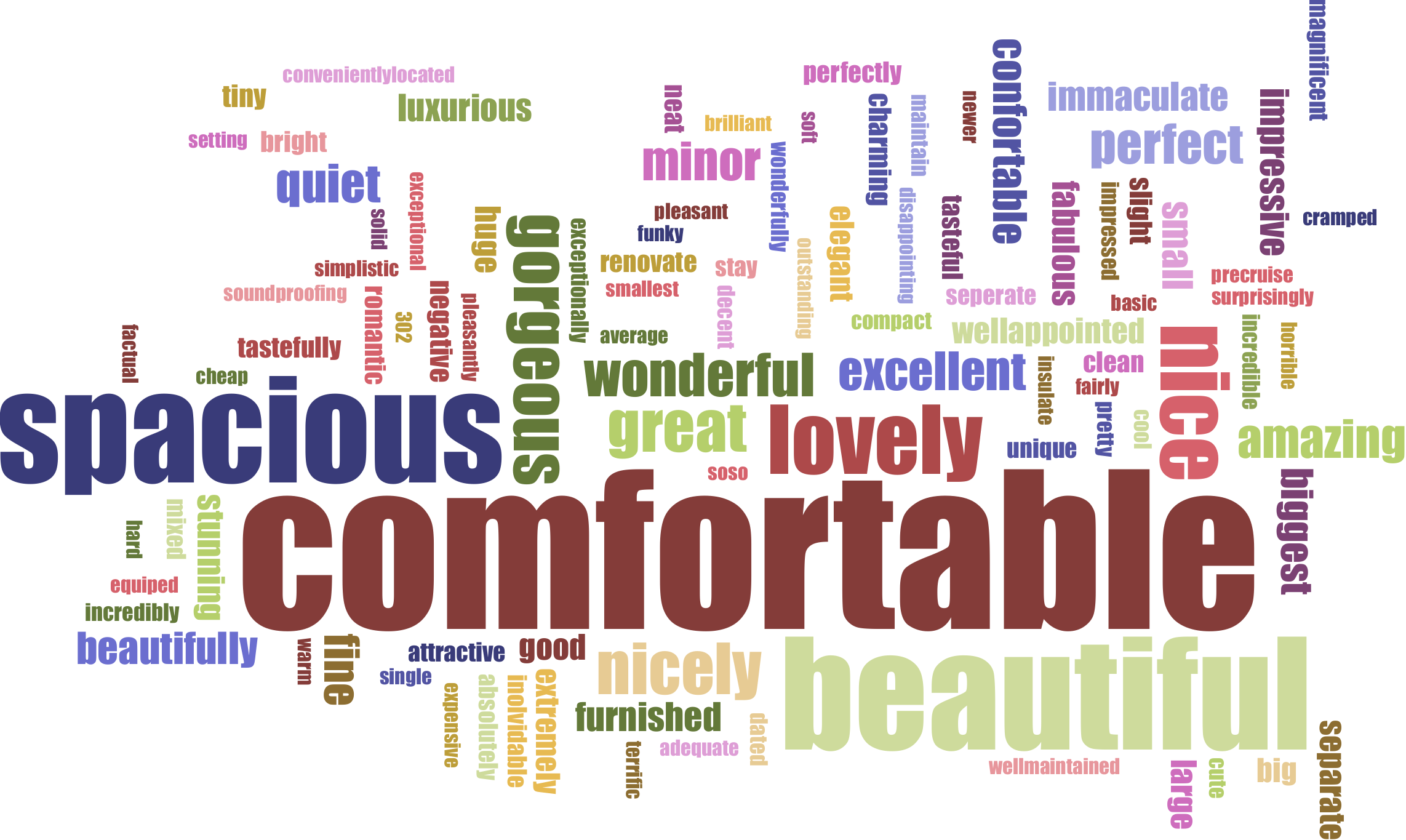}
        \caption{Room Positive}
        \end{subfigure}
        \begin{subfigure}[b]{0.24\linewidth}
        \includegraphics[width=\linewidth]{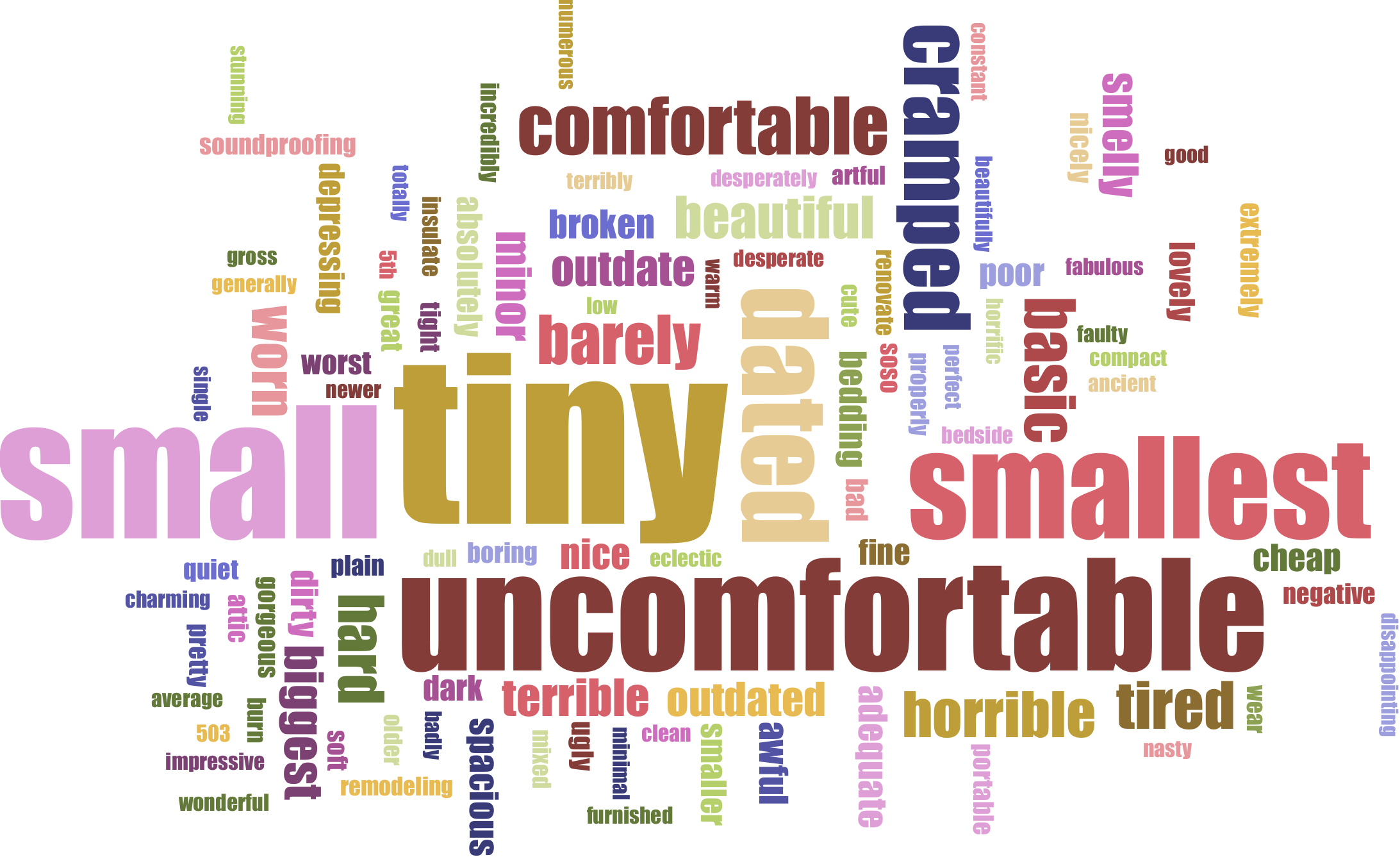}
        \caption{Room Negative}
        \end{subfigure}
        \begin{subfigure}[b]{0.24\linewidth}
        \includegraphics[width=\linewidth]{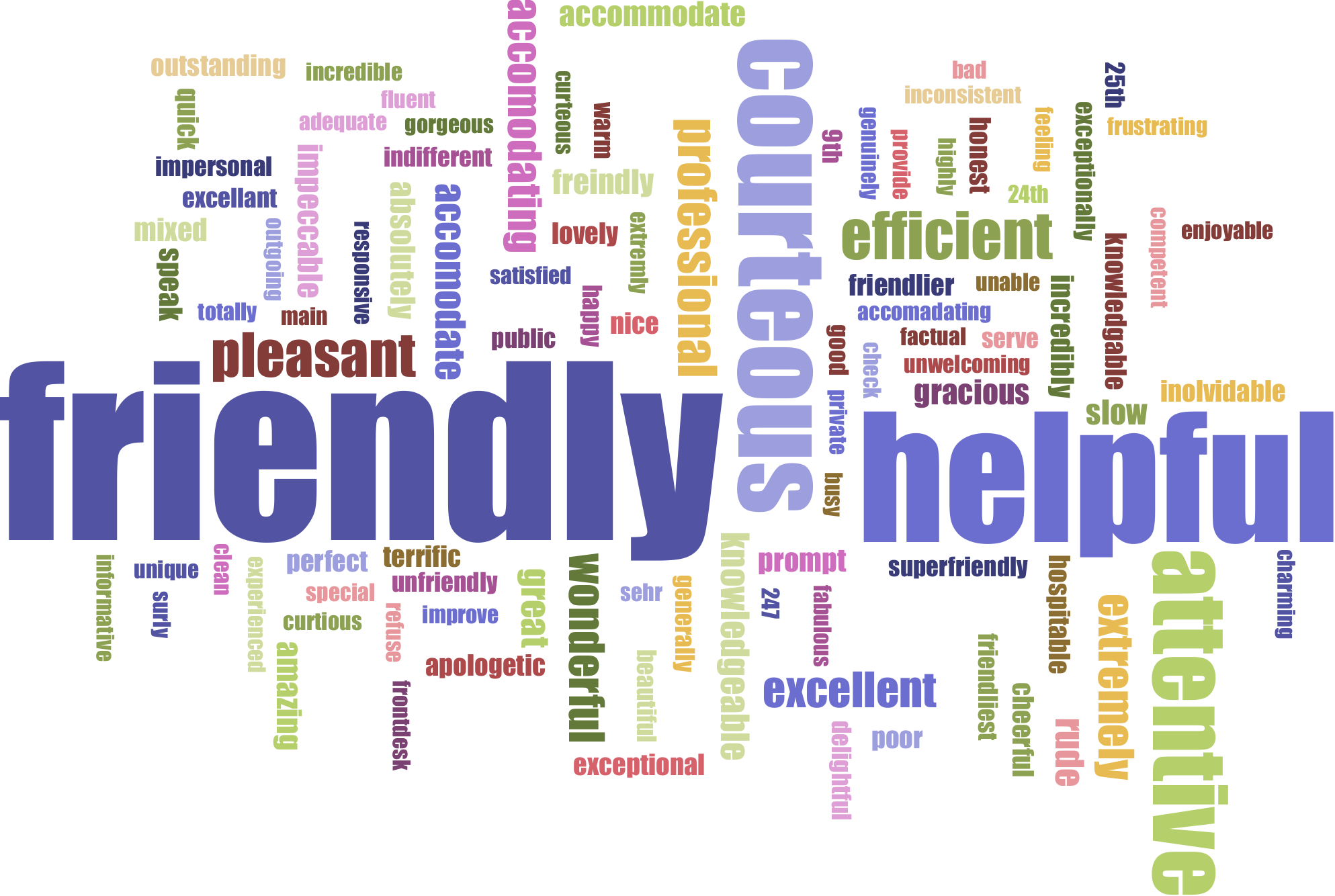}
        \caption{Service Positive}
        \end{subfigure}
        \begin{subfigure}[b]{0.24\linewidth}
        \includegraphics[width=\linewidth]{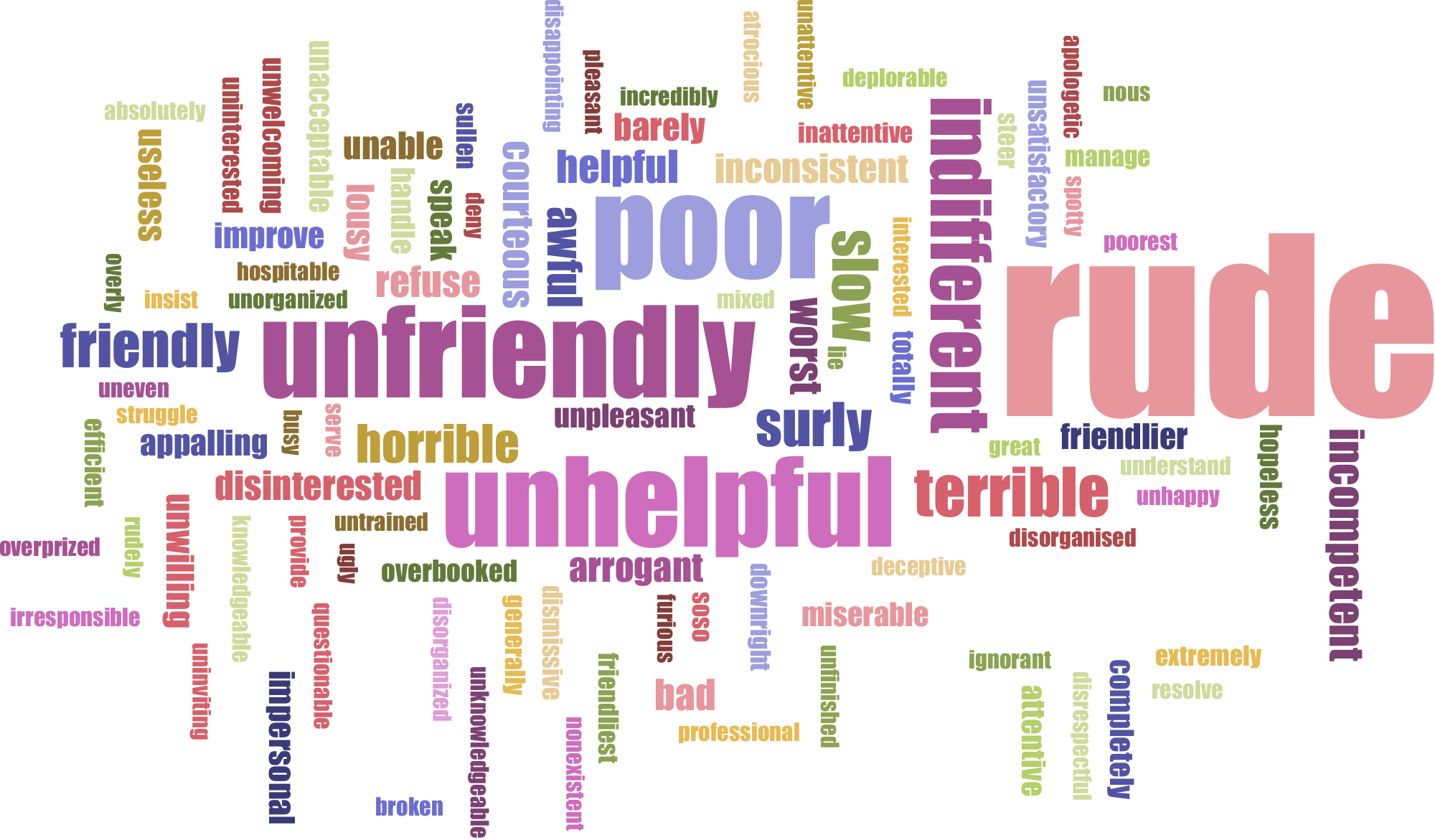}
        \caption{Service Negative}
        \end{subfigure}}
    
    \resizebox{1\linewidth}{!}{
        \begin{subfigure}[b]{0.24\linewidth}
        \includegraphics[width=\linewidth]{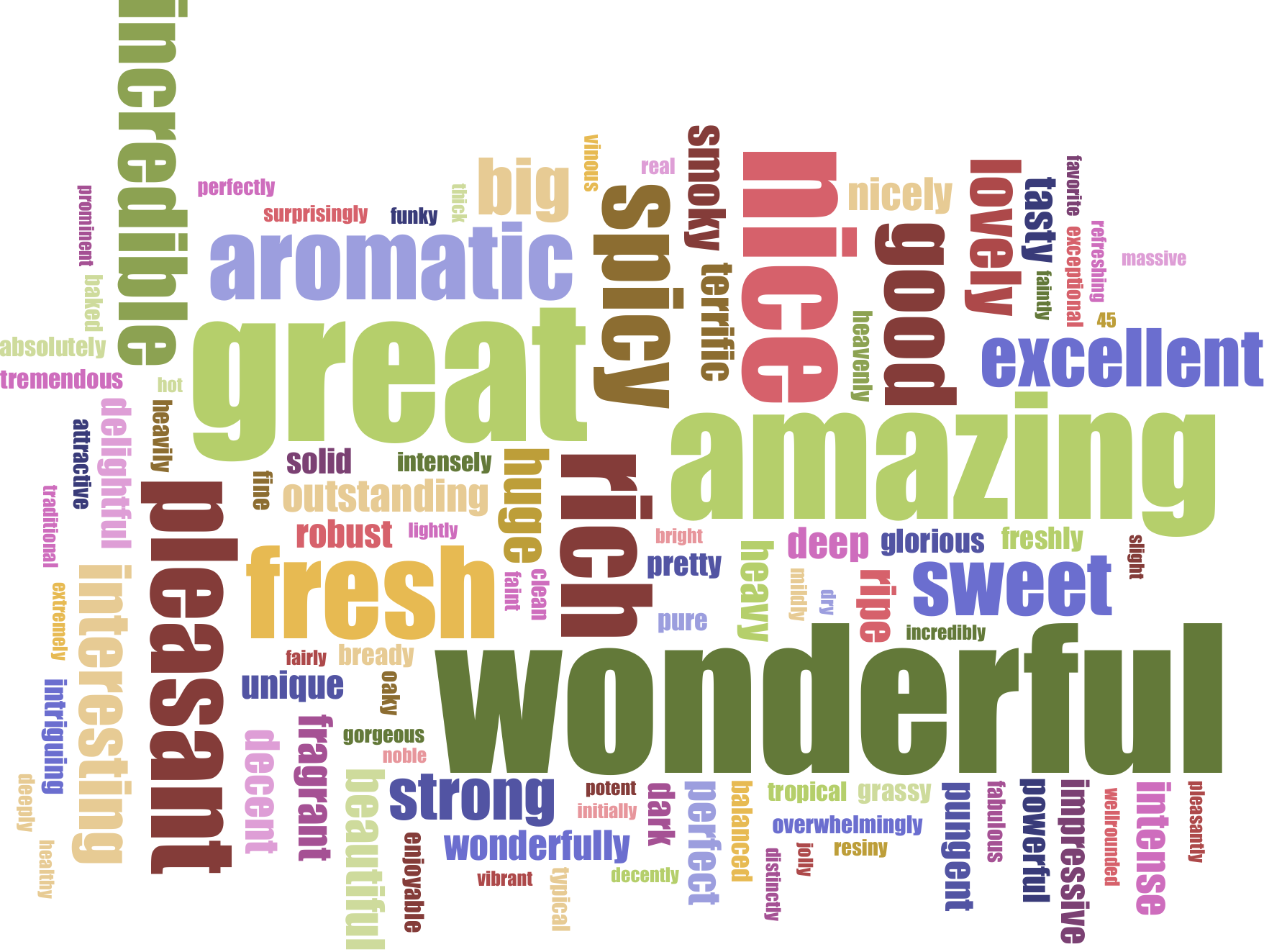}
        \caption{Smell Positive}
        \end{subfigure}
        \begin{subfigure}[b]{0.24\linewidth}
        \includegraphics[width=\linewidth]{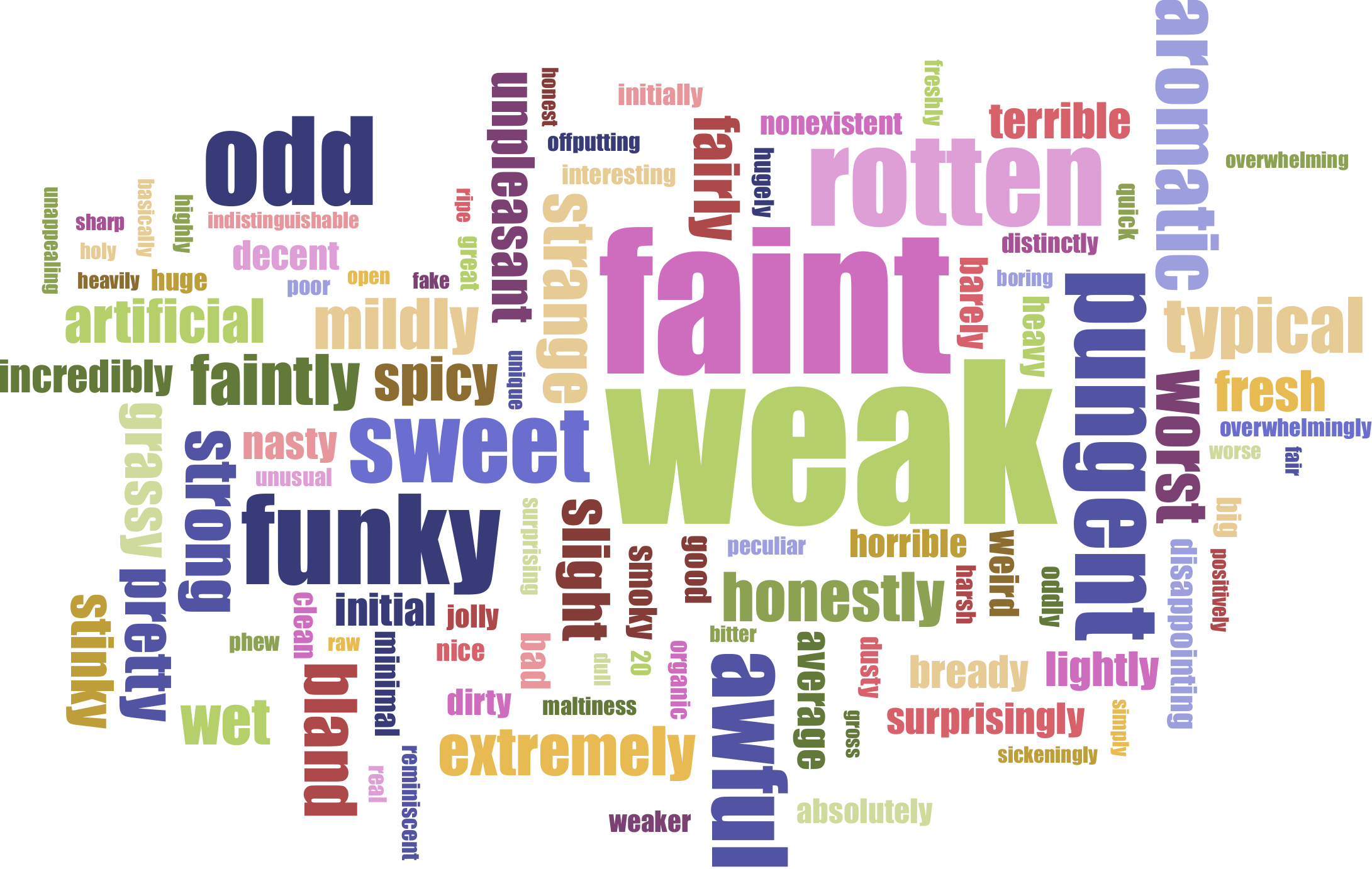}
        \caption{Smell Negative}
        \end{subfigure}
        \begin{subfigure}[b]{0.24\linewidth}
        \includegraphics[width=\linewidth]{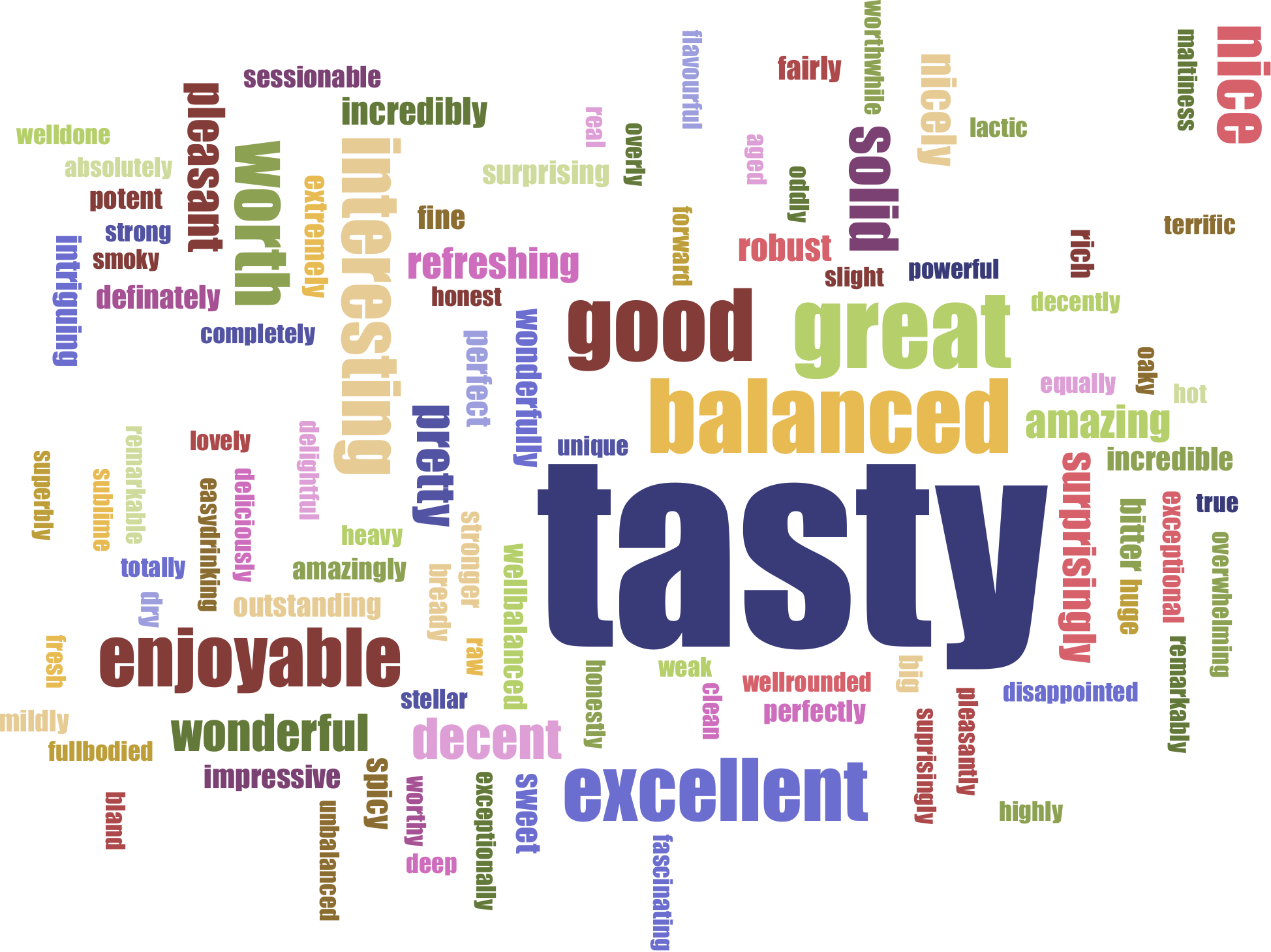}
        \caption{Taste Positive}
        \end{subfigure}
        \begin{subfigure}[b]{0.24\linewidth}
        \includegraphics[width=\linewidth]{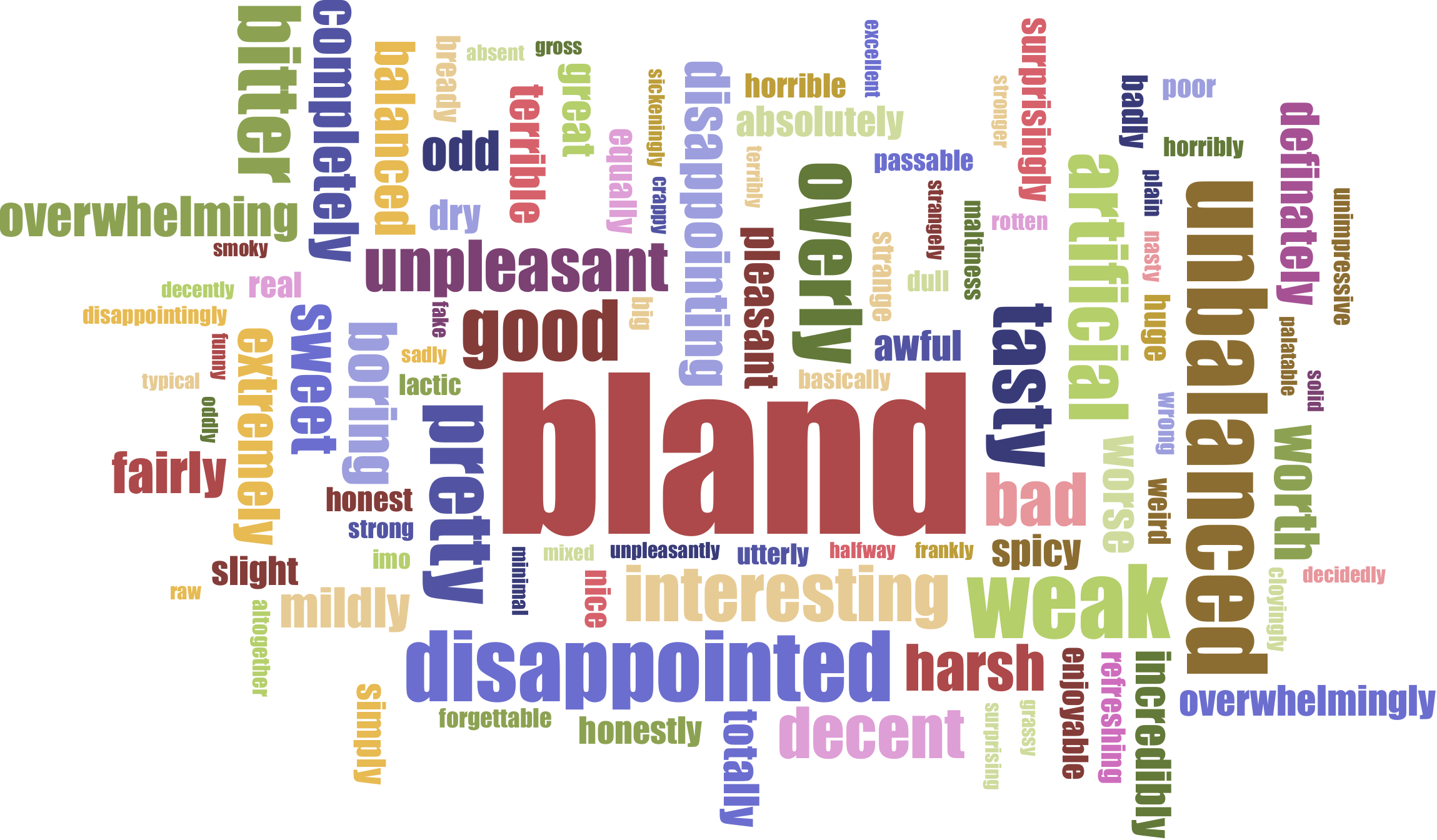}
        \caption{Taste Negative}
        \end{subfigure}}
        
    \resizebox{1\linewidth}{!}{
        \begin{subfigure}[b]{0.24\linewidth}
        \includegraphics[width=\linewidth]{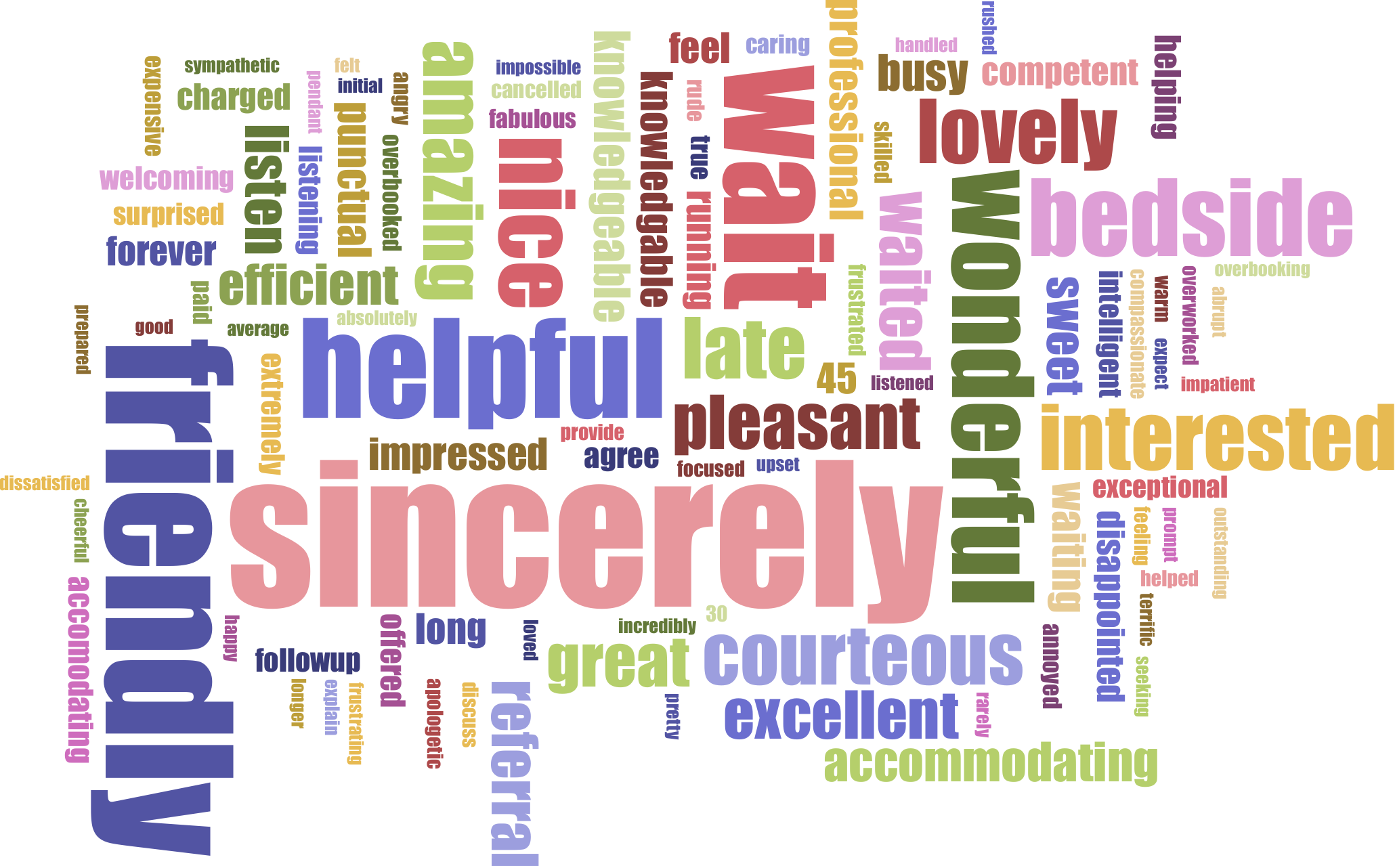}
        \caption{Staff Positive}
        \end{subfigure}
        \begin{subfigure}[b]{0.24\linewidth}
        \includegraphics[width=\linewidth]{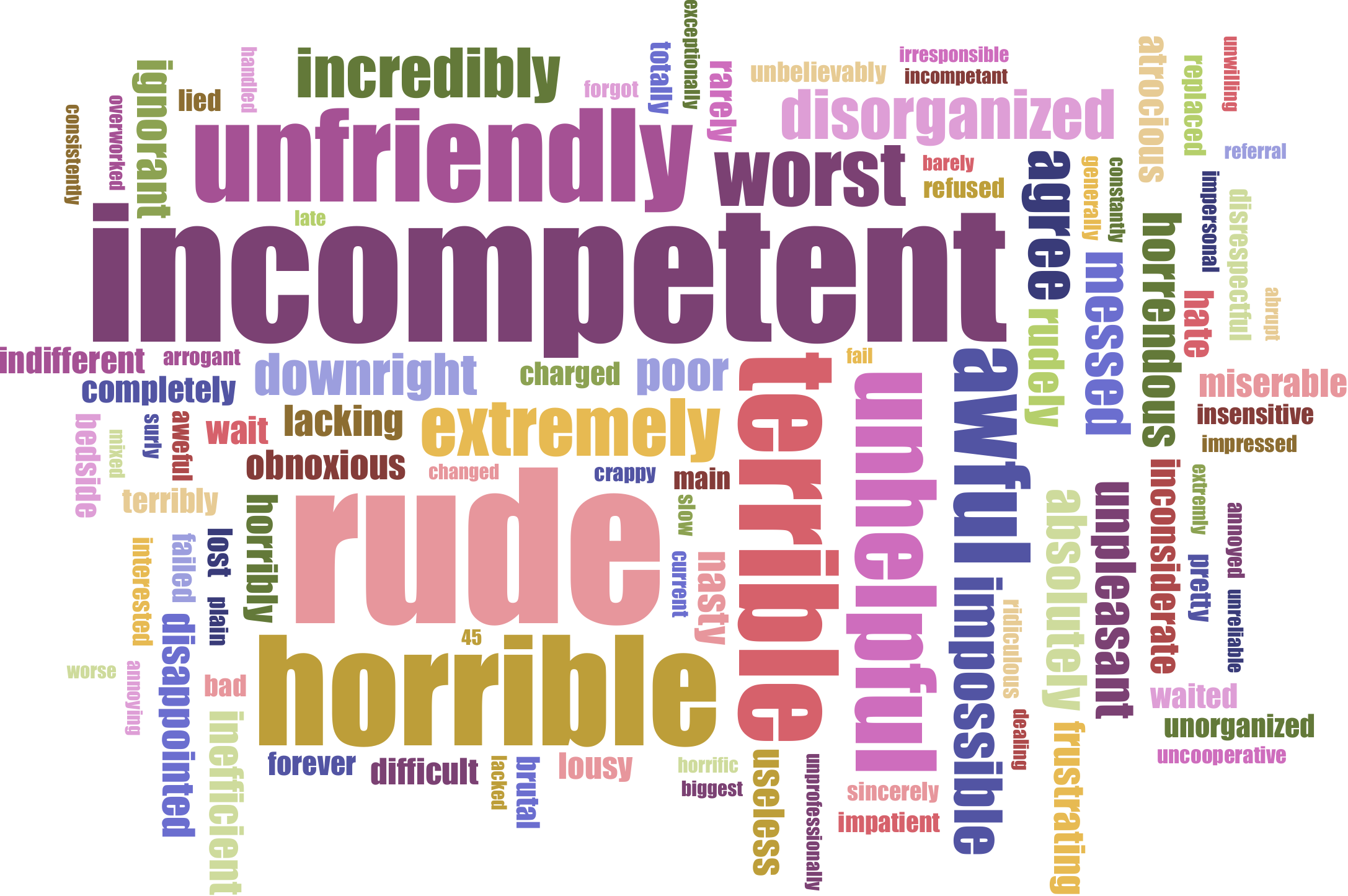}
        \caption{Staff Negative}
        \end{subfigure}
        \begin{subfigure}[b]{0.24\linewidth}
        \includegraphics[width=\linewidth]{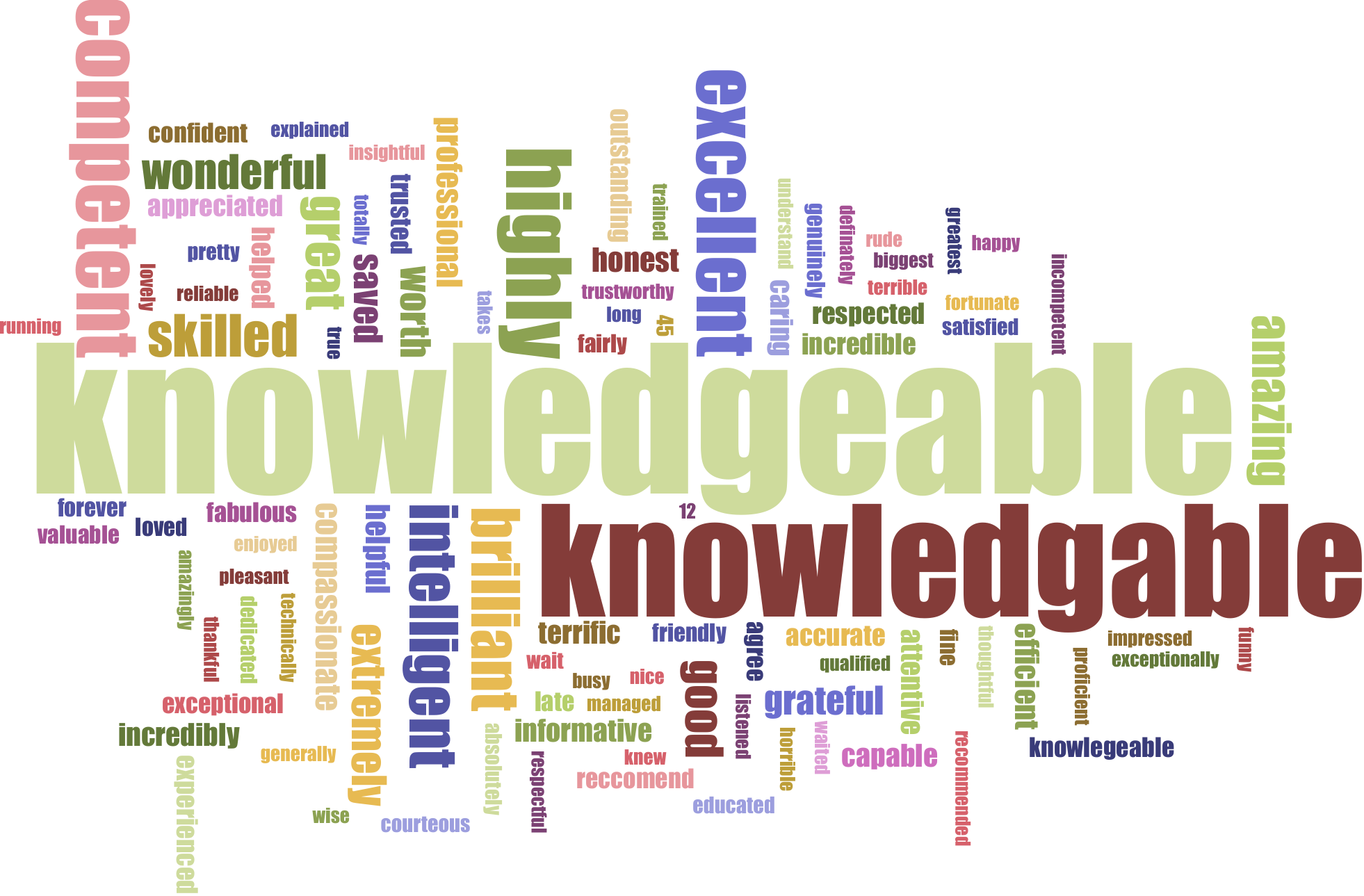}
        \caption{Knowledge Positive}
        \end{subfigure}
        \begin{subfigure}[b]{0.24\linewidth}
        \includegraphics[width=\linewidth]{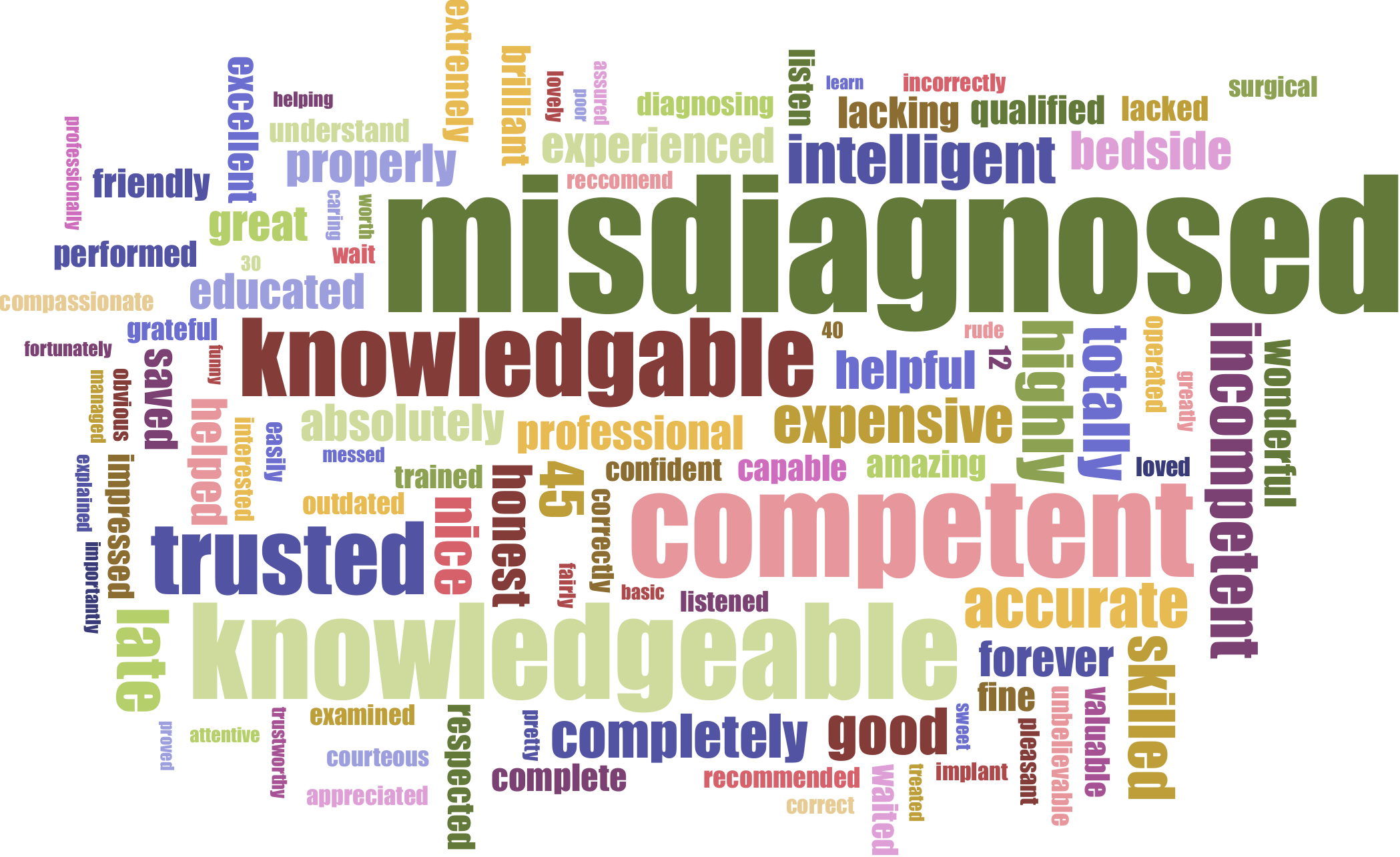}
        \caption{Knowledge Negative}
        \end{subfigure}}
        
    \caption{Word-cloud visualization of aspect-level opinion keywords for TripAdvisor-B (Top row), BeerAdvocate-B (Middle row) and RateMDs-B datasets (Bottom row).}
    \label{fig:word_cloud_sentiment_keywords}
\end{figure}

We also obtain aspect-specific opinion keywords from Trip-B, Beer-B, and RMD-B datasets, and show them in Fig.~\ref{fig:word_cloud_sentiment_keywords}.
From this figure (Top row), we observe that reviewers with positive experience usually live in ``\textit{comfortable, beautiful, spacious, lovely and gorgeous}'' rooms, and the staff are ``\textit{helpful, friendly, courteous and attentive}''.
While reviewers with negative experience may live in ``\textit{uncomfortable, small, cramped and tiny}'' rooms. Something may ``\textit{leak}'' and there are also problems with ``\textit{air conditioning}''.
The staff are ``\textit{rude, unhelpful and unfriendly}'' and the service is ``\textit{poor}''.
From Fig.~\ref{fig:word_cloud_sentiment_keywords} (Middle row), we learn that good beers should have ``\textit{great, amazing, wonderful, pleasant, aromatic, fresh, rich, and incredible}'' smell, and the taste may be ``\textit{tasty, great, balanced, enjoyable, and flavorful}''.
The smell of low-rated beers is ``\textit{faint, weak, pungent, odd, funky, and rotten}'', and the taste may be ``\textit{bland, unbalanced, disappointed, and sour}''.
From Fig.~\ref{fig:word_cloud_sentiment_keywords} (Bottom row), we find that good doctors usually have ``\textit{sincerely, friendly, helpful, and wonderful}'' staff and are ``\textit{knowledgeable, competent, intelligent, and excellent}''.
In a low-rated clinic, staff may be ``\textit{incompetent, rude, horrible, terrible, and unfriendly}'', and doctors may ``\textit{misdiagnose}'' conditions of patients and can be not ``\textit{competent, knowledgeable, or trusted}''.

From these figures, we can conclude that our deliberate self-attention mechanism is interpretable, and by leveraging our AKR method, it is a powerful knowledge discovery tool for online multi-aspect reviews, which answers research question \textbf{RQ4}.



\section{Conclusion}
\label{sec:conclusion}

In this paper, we proposed a multi-task deep learning model, namely FEDAR, for the problem of document-level multi-aspect sentiment classification.
Different from previous studies, our model does not require hand-crafted aspect-specific keywords to guide the attention and boost model performance for the task of sentiment classification.
Instead, our model relies on (a) a highway word embedding layer to transfer knowledge from pre-trained word vectors on a large corpus, (b) a sequential encoder layer whose output features are enriched by pooling and feature factorization techniques, and (c) a deliberate self-attention layer which maintains the interpretability of our model.
Experiments on various DMSC datasets have demonstrated the superior performance of our model. In addition, we also developed an Attention-driven Keywords Ranking (AKR) method, which can automatically discover aspect and opinion keywords from the review corpus based on attention weights.
Attention weights visualization and aspect/opinion keywords word-cloud visualization results have demonstrated the interpretability of our model and effectiveness of our AKR method.
Finally, we also proposed a LEcture-AuDience (LEAD) method to measure the uncertainty of deep neural networks, including our FEDAR model, in the context of multi-task learning.
Our experimental results on multiple real-world datasets  demonstrate the effectiveness of the proposed work.

\begin{acks}
This work was supported in part by the US National Science Foundation grants  IIS-1707498, IIS-1838730, and Amazon AWS credits.
\end{acks}

\bibliographystyle{ACM-Reference-Format}
\bibliography{ref}


\end{document}